\PassOptionsToPackage{capitalize}{cleveref}
\PassOptionsToPackage{hyphens}{url}
\documentclass{fairmeta}

\usepackage[utf8]{inputenc} % allow utf-8 input
\usepackage{url}
\usepackage{adjustbox}      % shrink wide tables to the text width
\usepackage{wrapfig}        % half-width figures beside the text

\usepackage{amsmath}
\usepackage{amssymb}
\usepackage{mathtools}
\usepackage{xspace}
\usepackage{enumitem}
\usepackage{tikz}
\usepackage{algorithm}
\usepackage[noend]{algpseudocode}
\makeatletter
\patchcmd{\ALG@doentity}{\item[]\nointerlineskip}{}{}{\PackageWarning{main}{algpseudocode noend patch failed}}
\makeatother
\usetikzlibrary{shapes.multipart, arrows.meta, positioning}

\makeatletter
\def\paragraph{\@startsection{paragraph}{4}{\z@}{0.6ex plus 0.3ex minus .2ex}{-1em}{\normalsize\bfseries}}
\makeatother

\theoremstyle{plain}

\theoremstyle{definition}

\theoremstyle{remark}

\usepackage[textsize=tiny]{todonotes}

\newcommand{\ours}{\texorpdfstring{\mbox{MIRIX-2.0}}{MIRIX-2.0}\xspace}

\providecommand{\ours}{MIRIX-2.0}
\makeatletter
\newcommand{\DeferredFloatBarrier}{\par\ifx\@deferlist\@empty\else\FloatBarrier\fi}
\makeatother

\title{\ours: Multi-Agent Memory System}

\author[*]{Wei-Chieh~Huang}
\author[*,\dagger]{Weizhi~Zhang}
\author[*]{Jiayan Nan}
\author[*]{Peng~Xia}
\author[*,\dagger]{Yu~Wang}

\affiliation{MIRIX AI}

\contribution[*]{Equal contribution}
\contribution[\dagger]{Corresponding author}
\abstract{
Long-term memory lets LLM agents personalize, stay consistent, and reuse what
worked before. Parametric memories need training, consolidation can lose details
for good, and keeping every record leaves a growing store to search. We present
\ours, a multi-agent memory system designed to be practical, accurate, and
efficient at once. It is the most accurate of the compared systems on four of five QA benchmarks.
It reaches 91.4\% on LoCoMo and 72.6\% on LongMemEval-S, where it leads the
strongest baseline by 14.3 points, and with AutoDream disabled it gains 31.0
and 22.0 points over HyperMem on single-hop and multi-hop document QA. As procedural memory, it yields the
highest mean success among the compared memory methods on ALFWorld and WebShop
and reaches 49.2\% on MetaClaw-Bench, against 44.0\% for MetaClaw's skill
library. Six typed memory agents keep complete records as the system of record,
and a provenance hypergraph indexes episodic and semantic records with n-ary
frames that cite their rows. Retrieval runs under explicit budgets, so the
evidence a question inspects does not grow with the history. AutoDream merges
duplicate anchors outside the write path without rewriting any record, and it
distills task sessions into skills. With AutoDream disabled, \ours answers
faster than HyperMem on all five timed benchmarks. It needs no fine-tuning, runs
with closed and open models, and its code is public.
}

\authoremails{\email{\{whuang80,wzhan42\}@uic.edu}, 
\email{njy@tongji.edu},
\email{pxia@cs.unc.edu},
\email{yuw164@ucsd.edu}}
\metadata[Website]{\url{https://mirix.io/}}
\metadata[GitHub]{\url{https://github.com/Mirix-AI/MIRIX}}

\begin{document}

\maketitle

% Keep the overview figure below the title box on the first page.
\suppressfloats[t]

\begin{figure*}[htpb]
    \centering
    \includegraphics[width=\textwidth]{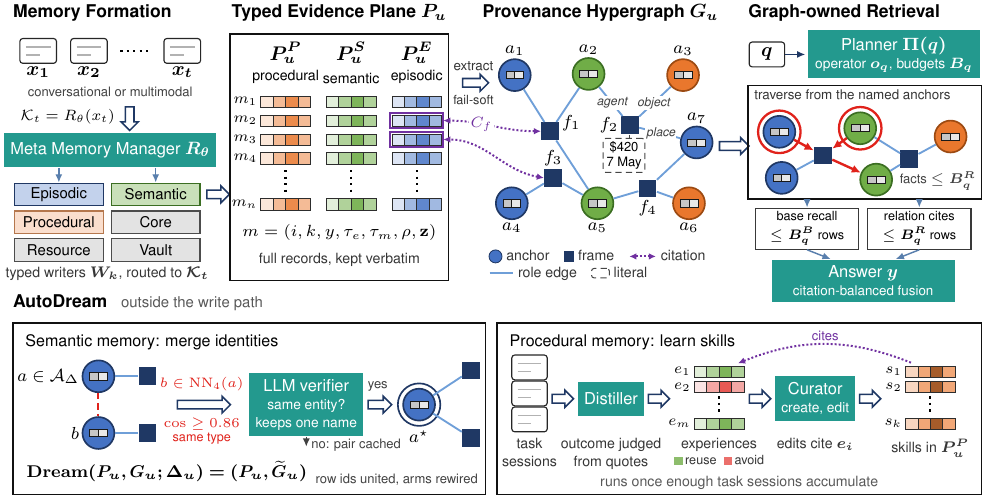}
    \caption{Overview of \ours. Typed writers fill the evidence plane $P_u$, the provenance
    hypergraph $G_u$ indexes it with n-ary frames that cite their sources, retrieval traverses
    $G_u$ under explicit budgets, and AutoDream consolidates outside the write path.}
    \label{fig:mirix-overview}
\end{figure*}

\section{Introduction}

Large language model (LLM) agents now work over horizons far longer than a
single context window~\citep{luo2025agentsurvey,agentmemorysurvey}. A personal
agent accumulates months of conversation with its user, and a task agent meets
similar problems across many episodes. Memory lets such agents carry
information across sessions~\citep{lscs}, so that they can personalize their
responses~\citep{zhang2026personaagent}, stay consistent with what they were
told~\citep{memgpt,mem0}, and reuse what worked
before~\citep{wang2025awm,xia2026skillrl}. Benchmarks now measure this
ability on long conversations~\citep{locomo,longmemeval}, long
documents~\citep{memoryagentbench}, multi-session agentic
tasks~\citep{he2026memoryarena}, and lifelong
personalization~\citep{memorycd}.

Benchmarks score a memory by its answers, but a team that deploys one asks more
of it. A memory that agents can rely on in practice has to meet three
requirements at once. It has to be accurate about the particulars of past
experience, such as a name, a date, a count, or the distinction between two
events, and about the connections between them. It has to be practical, which
means deployable with the models a team already uses, nothing to train, answers
that can be traced to stored records, and consolidation that cannot destroy what
was stored. It also has to be efficient, with answer latency and token cost per
question that stay bounded as the history grows. 
% Existing systems give up one of
% the three. Systems that consolidate records into extracted facts or
% abstractions~\citep{mem0,ye2026autodreamer} are compact, but the details they
% drop are gone for good, and consolidation can make a memory faulty even when it
% is distilled from useful experience~\citep{zhang2026faultymemories}. Graph
% memories organize the same content around entities and their
% relations~\citep{zep,edge2024graphrag}, yet binary triples split an n-ary fact
% into fragments~\citep{luo2025hypergraphrag,hypermem}.
% Systems that keep every record lose nothing, but they leave retrieval to search
% an ever larger and noisier history~\citep{agentmemorysurvey}. Parametric
% memories store experience in model weights or latent
% states~\citep{memoryllm,mplus,self-param,behrouz2024titans}, which requires
% training and rules out closed models.

Each family of existing systems gives up at least one of the three. Systems
that consolidate records into extracted facts or
abstractions~\citep{mem0,ye2026autodreamer} are compact but give up accuracy:
the details they discard are gone for good, and consolidation can make a
memory faulty even when it is distilled from useful
experience~\citep{zhang2026faultymemories}. Systems that keep every record and
retrieve from them directly lose nothing but give up efficiency, since
retrieval must search an ever larger and noisier
history~\citep{agentmemorysurvey}. Graph memories organize the same content
around entities and their relations~\citep{zep,edge2024graphrag}, but binary
triples split an n-ary fact into fragments and lose the connections among its
participants~\citep{luo2025hypergraphrag}. Hypergraph
memories~\citep{hypermem} group episodes under topics and facts under
episodes, but their hyperedges record which unit belongs where rather than
the participants of a fact, and retrieval descends from topic summaries
written at storage time, so a detail is reachable only after its topic has
been retrieved. File-system memories, such as the auto memory of Claude
Code~\citep{claudecodememory}, route retrieval through summaries in the same way: they keep notes in plain
files behind a size-capped index that is loaded into every session, and the
agent opens a file on the strength of its one-line summary. A detail that no
summary mentions is reachable only by string search, which neither resolves
an entity across files nor follows its relations, and index entries past the
cap are not loaded at the start of a session. Both families prune top down to
bound their cost, and both pay for the pruning with recall on details their
summaries do not reflect. Parametric memories store experience in model
weights or latent states~\citep{memoryllm,mplus,self-param,behrouz2024titans}
and give up practicality, since they require training and rule out closed
models.

\begin{figure}[!t]
    \centering
    \includegraphics[width=\linewidth]{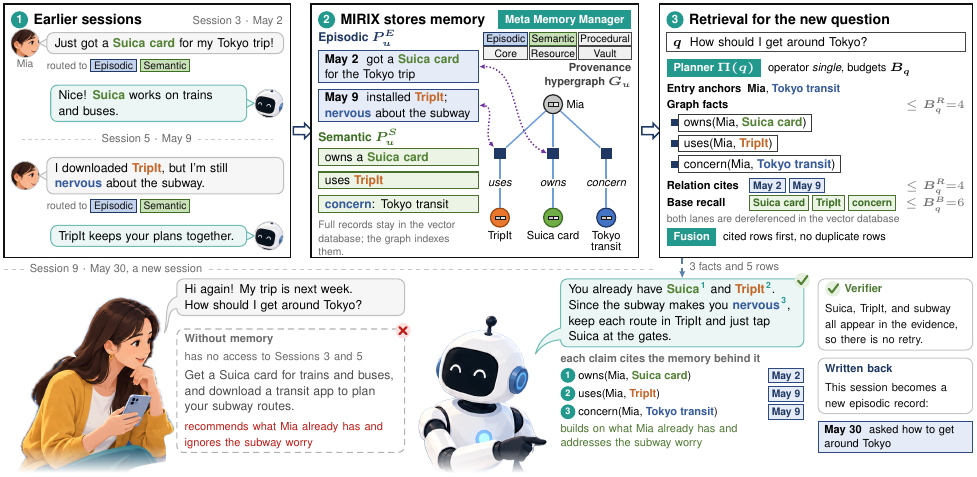}
    \caption{User-centric memory. Sessions are stored as episodic and semantic records, and a
    later question retrieves graph facts with their source rows to give a personalized answer.}
    \label{fig:mirix-user-centric}
\end{figure}

\begin{figure}[!t]
    \centering
    \includegraphics[width=\linewidth]{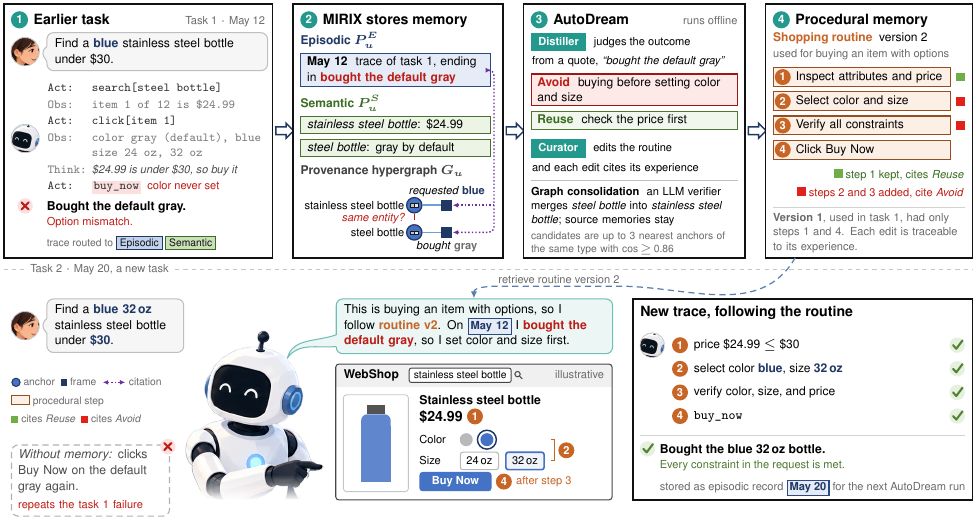}
    \caption{Agent-centric memory. AutoDream turns a failed purchase into experiences and a
    revised routine, which a later task follows to meet every constraint.}
    \label{fig:mirix-agent-centric}
\end{figure}

The requirements also weigh differently for the two uses of memory. Recent
studies separate user-centric memory, which personalizes an agent to its user,
from agent-centric memory, which accumulates the agent's own experience, and
they find that the two face reversed
challenges~\citep{agentmemorysurvey,harnessmemory}. A user-centric memory has
to record a user's history broadly and connect its pieces, because a later
question may depend on any detail of that history~\citep{memorycd,ma2026memaudit}.
An agent-centric memory has to be precise and refined instead. It distills
trajectories into strategies and
skills~\citep{zhao2024expel,ouyang2026reasoningbank,xia2026metaclaw}, and
retrieving too much can pull the agent's attention away from the context that
matters for its next action~\citep{harnessmemory}. A practical system has to
serve both uses from one store.
\Cref{fig:mirix-user-centric,fig:mirix-agent-centric} trace one example of each
setting through \ours.

In this work, we present \ours, a multi-agent memory system designed to be
practical, accurate, and efficient at once. It keeps the multi-agent design of
MIRIX~\citep{mirix} and rethinks how memory is stored, indexed, retrieved, and
consolidated so that a specific layer meets each requirement
(\Cref{fig:mirix-overview}). Six memory agents under a Meta Memory Manager write
complete records to a \emph{typed evidence plane} in a vector database, the system
of record, so details survive consolidation and every graph fact the answer
receives cites the rows it came from. The graph is only an index. The database write commits first, a failed graph write never rolls back a stored memory, and
a failed graph read falls back to direct evidence retrieval. A
\emph{provenance hypergraph} indexes the entities, events, relations, roles, and
times in episodic and semantic records as n-ary frames that cite the rows they
were extracted from and carry dates, prices, and counts as literals, and undated
events with similar wording keep distinct identities. At query time, a
deterministic planner sets explicit budgets, retrieval enters the hypergraph at the anchors that match the question, and the answer receives graph facts together
with the records they cite (\Cref{sec:retrieval}). The evidence a question inspects is therefore bounded by construction
and does not grow with the length of the history.

Like sleep-time and offline consolidation
methods~\citep{lin2025sleeptime,ye2026autodreamer}, AutoDream consolidates
memory outside the write path, and it treats the two kinds of memory in
opposite ways. For semantic memory, it consolidates identities and never
rewrites evidence. It pairs each anchor introduced by new semantic rows with its
nearest anchors of the same identity-stable type, such as a person or a place,
and asks a language model whether a pair denotes one entity. A confirmed merge
unites the citations of the two anchors and rewires the frames that mention
them, while event anchors, fact frames, and stored records stay unchanged, so a
merge can never destroy a stored memory. Its cost depends on the newly written
semantic frontier rather than on the total size of memory, verification runs
under explicit pair caps, and negative decisions are cached so that later cycles
do not pay for the same pair again. For procedural memory, AutoDream refines
instead. It judges each finished task session from quotes in its transcript,
distills lessons worth reusing and mistakes worth avoiding, and lets a
curator~\citep{zhang2026ace} create or edit skills whose edits can cite the
experiences behind them.

We evaluate \ours on conversational memory, long-document question answering,
personalized generation, and procedural memory for interactive agents. \ours
has the highest accuracy of all compared memory systems on both conversational
benchmarks, 91.4\% on LoCoMo in our runs and 72.6\% on LongMemEval-S, 14.3 points
above the strongest baseline. On LoCoMo, the hypergraph raises accuracy over flat
memory from 85.7\% to 88.4\%, with the largest gain on multi-hop questions, and
AutoDream raises every question type and lifts the overall score to 91.4\%. On
LongMemEval-S, the hypergraph alone adds 11.2 points. On the single-hop and multi-hop document QA tasks of
MemoryAgentBench~\citep{memoryagentbench}, with AutoDream disabled, it answers
83.0\% and 71.0\% of the questions, 31.0 and 22.0 points above the strongest
baseline, and on MemoryCD~\citep{memorycd} it has the highest value on all four
metrics for cross-domain review summarization. Used as procedural memory, \ours
gives the highest mean success among the compared memory methods on
ALFWorld~\citep{shridhar2021alfworld}, at 76.4\% with GPT-6 Luna against 73.6\%
for Agent Workflow Memory (AWM)~\citep{wang2025awm} and 69.4\% without memory, and on WebShop~\citep{yao2022webshop}, at
37.4\%. With GPT-5.2 it reaches 49.2\% accuracy on
MetaClaw-Bench~\citep{xia2026metaclaw}, against 44.0\% reported for the MetaClaw
skill library.

What makes these results usable is the cost at which they come. With AutoDream
disabled, \ours answers faster than HyperMem, the strongest baseline, on all five
benchmarks we timed, ingests the two conversational benchmarks in less time, and
uses fewer answer tokens on four of the five. As procedural memory, on WebShop it
uses the fewest training and test tokens among the memory methods while reaching
the highest success. The cost is paid when long documents are ingested. Building the graph makes
their ingestion about four times slower than flat memory and slower than that
baseline. The full system with AutoDream is not timed, and the retrieval bounds
cover the evidence a question inspects, not latency measured as the history
grows. No model is fine-tuned, the same system runs with closed and open
models, and the code is public.

Our contributions are as follows.
\begin{itemize}[leftmargin=*,nosep]
    \item \textbf{Accurate and practical.} A layered memory design in which six
    typed memory agents keep complete records as the system of record and a
    provenance-preserving n-ary hypergraph serves only as their index, with
    occurrence identities that keep distinct events apart and fail-soft
    indexing that never rolls back a stored memory.
    \item \textbf{One memory for users and agents.} To the best of our
    knowledge, \ours is the first memory system that handles both user-centric
    and agent-centric memory, serving personalization and the refinement of
    agent skills from one store.
    \item \textbf{Efficient and safe to consolidate.} Graph-owned retrieval runs
    under explicit budgets and returns graph facts together with the records
    that justify them. AutoDream merges duplicate entity anchors under explicit
    caps without touching stored memories, which adds 3.0 points on LoCoMo, and
    distills task sessions into skills for procedural memory.
\end{itemize}
The rest of the paper is organized as follows. \Cref{sec:method} presents
\ours. It states the problem and the three technical requirements, introduces
the typed evidence plane and the provenance hypergraph that indexes it,
describes graph-owned retrieval under explicit budgets, and explains how
AutoDream consolidates semantic and procedural memory outside the write path.
\Cref{sec:experiments} evaluates the accuracy of \ours on conversational recall,
long-document question answering, and personalized generation, its use as
agent memory on ALFWorld, WebShop, and MetaClaw-Bench, and its cost in
ingestion time, answer latency, and tokens. \Cref{sec:related} relates \ours to
prior work on memory systems for LLM agents, multi-agent systems, and memory
consolidation, and \Cref{sec:conclusion} concludes and discusses the
limitations of this study. \Cref{app:method} gives the details of the method,
and \Cref{app:results} gives the full results behind the figures.

\providecommand{\method}{\ours}

\section{Methodology}
\label{sec:method}

\method is designed to be practical, accurate, and efficient. It has to run
with the models a team already uses and leave stored records intact, it has to
keep a name, a date, a count, or the distinction between two events recoverable
after writing and after consolidation, and it has to answer from a bounded
evidence set however long the history grows. This section describes the
layered design that delivers the three.

\method separates storage from indexing (\Cref{fig:mirix-overview}). A \emph{typed evidence plane} stores complete
memories and is the system of record (\Cref{sec:typed-evidence}). A
\emph{provenance hypergraph} indexes the entities, events, relations, roles, and
times in those memories, and every fact in it cites the records it came from
(\Cref{sec:hypergraph}). Three paths operate on the two layers, and
\Cref{alg:mirix} states each of them in pseudocode. The write path stores a
memory and then indexes it. The read path plans a query, traverses the
graph, and dereferences only the records it selects (\Cref{sec:retrieval}).
AutoDream, the third path, runs outside the other two (\Cref{sec:autodream}). On semantic memory it
merges duplicate anchors in the hypergraph and returns the evidence store
unchanged, so semantic consolidation changes how evidence is connected, never
what evidence exists. On procedural
memory, AutoDream instead distills task sessions into skills that cite the
experiences behind them. \Cref{app:method-invariants} collects the properties that hold regardless of
how the language models behave.

\subsection{Problem setup}
\label{sec:method-setup}

For a user \(u\), the system observes a stream
\(\mathcal X_u=(x_1,\ldots,x_T)\) of conversational or multimodal inputs. When it
writes \(x_t\), future queries are unknown, so the writer must produce a
persistent state
\begin{equation}
    \mathcal S_u^t=(P_u^t,G_u^t)
\end{equation}
from which a later query \(q\) can be answered with a bounded evidence set
\(\mathcal R(q;\mathcal S_u^t)\). The evidence store \(P_u\) collects the typed
stores, among them the episodic store \(P_u^E\), the semantic store \(P_u^S\),
and the procedural store \(P_u^P\). We seek high answer utility under three
technical requirements.
\begin{enumerate}[leftmargin=*,nosep]
    \item \textbf{Source fidelity.} Concrete evidence remains recoverable after
    writing and after consolidation.
    \item \textbf{Bounded access.} A query inspects a small graph neighborhood
    and a bounded set of cited rows instead of scanning the full history.
    \item \textbf{Temporal identity.} Repeated descriptions of one event may
    share evidence, but distinct occurrences must not collapse merely because
    they are worded alike.
\end{enumerate}
Source fidelity and temporal identity are what make the system accurate and
its answers traceable, and bounded access is what keeps it efficient. The rest
of this section shows how the two layers meet these requirements.
Each subsection explains a block of \Cref{alg:mirix}, and the numbers on its right
point to the equations that define each step.

\begin{algorithm}[t]
\caption{\method for one user \(u\). Calls marked \(\star\) use a language model.}
\label{alg:mirix}
\small
\algrenewcommand{\algorithmiccomment}[1]{\hfill#1}
\begin{minipage}[t]{0.485\textwidth}
\begin{algorithmic}[1]
\Procedure{Write}{$x_t$}
  \For{\(k\in R_\theta^\star(x_t)\)}\label{alg:route}
    \State \(m\gets W_k^\star(x_t,P_u)\), \(\textsc{Commit}(P_u,m)\)\Comment{\eqref{eq:memory-record}}\label{alg:commit}
    \If{\(k\in\{\mathrm{episodic},\mathrm{semantic}\}\)}
      \State \(\Phi\gets\textsc{Extract}^\star(m)\cup\textsc{Repair}^\star(m)\)\Comment{\eqref{eq:frame}}\label{alg:extract}
      \State \(A_m\gets\textsc{Admit}(\textsc{Canon}^\star(\Phi))\)
      \State \(I_a\gets I_a\cup\{i_m\}\) for \(a\in A_m\)
      \For{\(f\in\Phi\) with \(|A_f|\ge 2\)}
        \State \(\textsc{Upsert}(G_u,\operatorname{id}(f),f)\)\Comment{\eqref{eq:semantic-frame-id}, \eqref{eq:episodic-frame-id}}\label{alg:upsert}
        \State \(C_f\gets C_f\cup\{i_m\}\)
      \EndFor
    \EndIf
  \EndFor
\EndProcedure
\Procedure{Read}{$q$}
  \State \(\Pi\gets\textsc{Plan}(q)\)\Comment{\eqref{eq:query-plan}}\label{alg:plan}
  \State \(A_q\gets\textsc{Top}(\textsc{Recall}(q),\,s_A,\,K_A)\)\Comment{\eqref{eq:anchor-score}}
  \State \(F_q\gets\textsc{Top}(\textsc{Adj}(A_q,\mathcal P_q),\,s_F,\,B_q^R)\)\Comment{\eqref{eq:fact-score}}
  \State \Return \(F_q\), \(\textsc{Fuse}(\textsc{Rows}(A_q),\,\bigcup_{f\in F_q}C_f;\,B_q^B,B_q^R)\)
\EndProcedure
\algstore{mirixalg}
\end{algorithmic}
\end{minipage}\hfill
\begin{minipage}[t]{0.485\textwidth}
\begin{algorithmic}[1]
\algrestore{mirixalg}
\Procedure{Answer}{$q$}
  \State \(\hat y\gets\textsc{Agent}^\star(q)\) with at most five \textsc{Read} calls
  \If{\(\textsc{Check}(\hat y)\)}
    \State \(\hat y\gets\textsc{Rewrite}^\star(\hat y,\textsc{Read}(q'))\)\label{alg:retry}
  \EndIf
  \State \Return \(\hat y\)
\EndProcedure
\Procedure{SemanticDream}{$\Delta_u$}\Comment{\eqref{eq:dream-contract}}
  \For{\((a,b)\in\mathcal C_\Delta\setminus\mathcal N\)}\Comment{\eqref{eq:dream-candidates}}\label{alg:dream-pairs}
    \If{\(\textsc{Same}^\star(a,b)\)}
      \State \(\textsc{Merge}(a\to b)\)
    \Else
      \State \(\mathcal N\gets\mathcal N\cup\{(a,b)\}\)
    \EndIf
  \EndFor
\EndProcedure
\Procedure{ProceduralDream}{$S_u$}
  \State \(e_{1:m}\gets\textsc{Distill}^\star(S_u)\)
  \State \(P_u^P\gets\textsc{Curate}^\star(P_u^P,e_{1:m})\)\label{alg:procedural}
\EndProcedure
\end{algorithmic}
\end{minipage}
\end{algorithm}

\subsection{Typed evidence plane}
\label{sec:typed-evidence}

\paragraph{Multi-agent routing.}
A Meta Memory Manager routes each input \(x_t\) to the subset
\(\mathcal K_t=R_\theta(x_t)\) of memory types it updates (\Cref{alg:mirix},
line~\ref{alg:route}). Each of the six types, Core, Episodic, Semantic,
Procedural, Resource, and Knowledge Vault, has its own writer, and the writer
of a selected type \(k\) produces the record \(m_t^{(k)}=W_k(x_t,P_u^{t-1})\)
and owns the schema and update operations of that type. Only episodic and
semantic records are indexed by the hypergraph. \Cref{app:method-routing} gives
the roles of the six types and the reason for this split.

\paragraph{Evidence records.}
Each of the six memory types keeps its records in its own schema. The hypergraph
indexes the episodic and semantic records, and each of them is represented as
\begin{equation}
    m=(i,k,y,\tau_e,\tau_m,\rho,\mathbf z),
    \label{eq:memory-record}
\end{equation}
where \(i\) is an immutable row identifier,
\(k\in\{\mathrm{episodic},\mathrm{semantic}\}\) is its memory type, \(y\) is the
stored text, \(\tau_e\) is the event time when one applies, \(\tau_m\) is the
mention or write time, \(\rho\) records source and chunk provenance, and
\(\mathbf z\) holds the search embeddings. Full summaries and details remain in
the vector database. The graph stores only identifiers and compact structural metadata.

Writers keep concrete details such as names, counts, and dates in each record,
and a record is committed to the vector database before graph construction
begins (line~\ref{alg:commit}), so a failed graph write never rolls back a
stored memory (\Cref{app:method-routing}).

\subsection{A provenance-preserving n-ary hypergraph}
\label{sec:hypergraph}

The second half of the write path turns each episodic or semantic record into
graph structure (\Cref{alg:mirix}, lines~\ref{alg:extract}--\ref{alg:upsert}).
For each user, \method maintains a hypergraph
\begin{equation}
    G_u=(\mathcal A_u,\mathcal F_u,\mathcal E_u)
\end{equation}
with entity or concept anchors \(\mathcal A_u\), reified fact frames
\(\mathcal F_u\), and role-typed incidence edges \(\mathcal E_u\).
\Cref{app:method-implementation} shows the hypergraph \method builds for one LoCoMo conversation.

\paragraph{Anchors.}
An anchor
\begin{equation}
    a=(i_a,n_a,t_a,\mathbf e_a,I_a^E,I_a^S)
\end{equation}
stores a canonical name \(n_a\), an entity type \(t_a\), a name embedding
\(\mathbf e_a\), and the identifiers \(I_a^E\) and \(I_a^S\) of the episodic and
semantic rows that mention it. A deterministic key \(\kappa(n)\) normalizes
case, punctuation, function words, word order, and singular and plural variants
of a name while keeping numbers, and a deterministic admission gate rejects
generic filing terms and dialogue-role pseudo-entities and caps the anchors
admitted from one record (\Cref{app:method-graph}).

\paragraph{Registry-guided canonicalization.}
Surface normalization cannot decide whether two semantically close names denote
the same entity. For every extracted entity, approximate nearest-neighbor search
therefore retrieves a few existing anchors of the user above a similarity
threshold, and one language-model call per record makes a conservative
same-entity decision for all of its entities. A rename is accepted only when it
selects a retrieved candidate, so embeddings propose comparisons but never
authorize a merge on their own. The neighbor count and the threshold, with their relation to the vector-index score, are in \Cref{app:method-graph}.

\paragraph{N-ary frames.}
For each episodic or semantic record, an extractor produces a set of frames
\begin{equation}
    f=
    \left(
        p,\,
        \{(r_j,a_j)\}_{j=1}^{d_f},\,
        \ell_f,\,
        \tau_e,\tau_m,\,
        C_f
    \right),
    \label{eq:frame}
\end{equation}
where \(p\) is a canonical predicate, \(r_j\) is the open-vocabulary role of
anchor \(a_j\), \(\ell_f\) maps literal attributes to their values, and \(C_f\)
is the set of identifiers of the vector-database records the frame cites. A flight, for example,
becomes one frame with agent, origin, destination, carrier, and companion arms
instead of several disconnected triples, and owns(Mia, Suica card) in
\Cref{fig:mirix-user-centric} is a frame that cites the May 2 record. The extractor is instructed to keep dates, prices, counts, durations, and
measurements as literals on the frame rather than as anchors, so that a value
stays attached to the assertion it qualifies. A frame is materialized only if
its arguments resolve to at least two distinct anchors of the user. When a deterministic audit finds that
the extraction misses high-signal source details, one bounded repair call may
add frames that the source supports, and it cannot remove anything the primary
extraction produced. \Cref{app:method-graph} describes the audit.

\paragraph{Occurrence identity.}
Semantic claims and episodic occurrences follow different identity rules. Let
\(\sigma(f)\) be the sorted set of role and anchor keys
\(\{r_j:\kappa(n_{a_j})\}\). A semantic frame is identified by
\begin{equation}
    \operatorname{id}_S(f)
    =H(u,p,\sigma(f)),
    \label{eq:semantic-frame-id}
\end{equation}
so repeated support for the same durable claim accumulates citations on one
node. An episodic frame adds an occurrence key,
\begin{align}
    \operatorname{id}_E(f)
        &=H(u,p,\sigma(f),\chi(f)), \label{eq:episodic-frame-id}\\
    \chi(f)
        &=
        \begin{cases}
        \text{explicit event time}, & \text{if extracted},\\
        \text{immutable memory id}, & \text{otherwise}.
        \end{cases}
\end{align}
Two mentions of the same dated event therefore share one frame, while two undated
events with similar wording, such as two runs on different days, stay separate. A
frame keeps the time of the event apart from the time it was mentioned, and whether
the event is planned or done does not change its identity.

\subsection{Graph-owned evidence retrieval}
\label{sec:retrieval}

\method treats retrieval as the construction of a bounded evidence set rather
than as open-ended similarity search. Episodic and semantic candidates must
enter through the graph, and the vector database dereferences and reranks only the
identifiers that graph traversal selects. This boundary keeps a hidden
full-table search from doing the work attributed to the graph. Only a graph
read that raises an error falls back to direct evidence retrieval. A query passes
through planning, anchor entry, expansion along two evidence lanes, and fusion,
and a verifier then checks the answer
(\Cref{alg:mirix}, lines~\ref{alg:plan}--\ref{alg:retry}). For the question in
\Cref{fig:mirix-user-centric}, the planner chooses the single operator, enters
the graph at Mia and Tokyo transit, and returns three facts with five rows.

\paragraph{Query plan.}
A deterministic planner maps a query to
\begin{equation}
    \Pi(q)=
    \left(
        o_q,\mathcal P_q,\mathcal C_q,\mathcal Y_q,\mathcal T_q,
        B_q^B,B_q^R
    \right),
    \label{eq:query-plan}
\end{equation}
where the operator \(o_q\) is single, temporal, intersection, list-union, or
count, \(\mathcal P_q\) holds canonical relation families and their surface
forms, \(\mathcal C_q\) holds non-relational constraints, \(\mathcal Y_q\) holds
expected answer types, and \(\mathcal T_q\) holds date and temporal-order
constraints. The two budgets cap the base-recall rows and the relation-cited
rows, and \(B_q^R\) also caps the graph facts that the relation lane renders.
The budgets grow with the operator, from 6 base rows and 4 cited rows for a
single-value lookup to 8 and 12 for a count, since counting and set
construction need broader evidence.
\Cref{tab:lane-budgets} in \Cref{app:method-retrieval} lists all five
operators.

\paragraph{Anchor entry.}
The query embedding first recalls fifteen anchors by name. A relation query adds
the anchors that the question names exactly, and the union is scored by
\begin{equation}
    s_A(a,q)=s_{\mathrm{vec}}+s_{\mathrm{exact}}+
    s_{\mathrm{lex}}+s_{\mathrm{pred}}+s_{\mathrm{spec}}
    -s_{\mathrm{hub}}.
    \label{eq:anchor-score}
\end{equation}
The vector score dominates. Exact mention, token overlap, predicate
compatibility, and multi-token specificity break close ties, and a logarithmic
degree penalty suppresses generic hubs. The top fifteen anchors are kept, which
is the cap \(K_A\) of \Cref{alg:mirix}. The cap on exactly named anchors and the hub and predicate rules are in \Cref{app:method-retrieval}.

\paragraph{Two evidence lanes.}
The selected anchors open two bounded lanes. In the base-recall lane, anchors
yield the episodic and semantic identifiers directly associated with them, and
traversing an incident n-ary frame adds the frame's cited memories and a bounded
set of memories associated with its co-participants. In the relation-citation
lane, which opens when the query plan contains a relation, the exactly named anchors seed an indexed adjacency lookup over their incident
frames. Predicate filtering applies before a cap of 240 frames, the fact score
below ranks the rest, and at most \(B_q^R\) of them are rendered as explicit
graph facts that contribute their exact citations into the vector database. The provenance
recomputation of the base lane and the object filter of the relation lane are
in \Cref{app:method-retrieval}.

\paragraph{Fact scoring.}
Each relation-lane frame receives the additive score
\begin{align}
    s_F(f,q)={}&
    s_{\mathrm{rel}}+s_{\mathrm{entity}}+s_{\mathrm{constraint}}
    +s_{\mathrm{type}} \notag\\
    &+s_{\mathrm{time}}+s_{\mathrm{intersection}}
    +s_{\mathrm{specificity}} \notag\\
    &+s_{\mathrm{role}}+s_{\mathrm{state}}.
    \label{eq:fact-score}
\end{align}
The first seven terms score relation, named-entity, literal, answer-type, and
temporal compatibility, a shared object for intersection questions, and the
number of arguments a frame binds. The role term \(s_{\mathrm{role}}\) rewards agreement
between the argument direction implied by the question and the frame's typed
roles, and it penalizes incompatible roles. The state term \(s_{\mathrm{state}}\)
rewards an event time that matches a requested date and penalizes planned
occurrences in count, list-union, and temporal questions.
\Cref{app:method-retrieval} describes its fallback and the treatment of intersection questions.

\paragraph{Citation-balanced fusion.}
Rows in each lane are reranked, and fusion takes up to \(B_q^R\) cited rows,
then up to \(B_q^B\) base rows, and fills the remaining capacity from the rest
of either lane without duplicating row identifiers. The returned evidence
therefore contains both the graph facts and the source rows that justify them.
The reranking prior, the per-type split, and the rule for near-duplicate paraphrases are also in \Cref{app:method-retrieval}.

\paragraph{Answer-time verification.}
In our harness, the answering agent may call the retriever up to five times.
After it drafts an answer, a rule-based check requests one more search and one
rewrite by the answering model when the draft shows explicit uncertainty, lacks a value the question asks for,
or disagrees with the evidence ledger, for example a count without a number or a
month that neither the evidence nor the question contains. It also fires on
every intersection question (line~\ref{alg:retry}). The check runs in the answering agent rather than in
the memory server, and it never inspects a reference answer.
\Cref{app:method-verification} lists every trigger and the query of the extra
search.

\subsection{AutoDream: consolidation outside the write path}
\label{sec:autodream}

AutoDream runs outside the critical write path
(\Cref{alg:mirix}, lines~\ref{alg:dream-pairs}--\ref{alg:procedural}). It treats
the two kinds of memory it consolidates in opposite ways. On semantic memory it
is an explicit service call that never schedules itself, and it refines
identities under the contract
\begin{equation}
    \operatorname{Dream}(P_u,G_u;\Delta_u)
    = (P_u,\widetilde G_u),
    \label{eq:dream-contract}
\end{equation}
where \(\Delta_u\) is the newly written semantic frontier and \(\widetilde G_u\)
is the consolidated hypergraph. The evidence store \(P_u\) is returned
unchanged. On procedural memory it starts in the background once five task
sessions are pending, and it turns them into skills.

\paragraph{Semantic frontier.}
Each call receives the exact source-chunk identifiers of the batch just
written, which resolve to the newly written semantic rows. These rows define a
small frontier \(\Delta_u\), so the cost of consolidation depends on new
activity rather than on the total size of memory. The invocation schedule of our protocol is in \Cref{app:method-autodream}.

\paragraph{Candidate generation.}
Only anchors that are backed by semantic rows and belong to an identity-stable
type \(t\in\mathcal T_{\mathrm{safe}}=\{\text{person, organization, location,
object, concept}\}\) are eligible. For each eligible anchor \(a\in\mathcal A_\Delta\) backed by a frontier row, AutoDream takes the four nearest entries of the name index, which include \(a\) itself,
and forms the candidate pairs
\begin{equation}
    \mathcal C_\Delta=
    \left\{
    (a,b):
    \begin{array}{l}
    a\in\mathcal A_\Delta,\;
    b\in\operatorname{NN}_4(a)\setminus\{a\},\\
    \cos(\mathbf e_a,\mathbf e_b)\ge 0.86,\\
    t_a=t_b\in\mathcal T_{\mathrm{safe}}
    \end{array}
    \right\}.
    \label{eq:dream-candidates}
\end{equation}
Episodic-only anchors are excluded by construction. An LLM decides whether each
pair denotes the same entity and, if it does, chooses one of the two existing
names as the canonical representative. Its output is constrained to the
submitted pair, and invented representatives are rejected. In
\Cref{fig:mirix-agent-centric}, \emph{steel bottle} and \emph{stainless steel
bottle} form such a pair.

\paragraph{Bounded schedule.}
Every Dream round verifies at most a fixed number of unseen pairs, and the
final round also considers the eligible anchors that no earlier round used as a
source. Explicit negative decisions are cached as the set \(\mathcal N\) of
\Cref{alg:mirix}, so later cycles do not pay again to reconsider the same pair.
\Cref{app:method-autodream} states the caps of the intermediate, final, and
repair rounds and the cache key.

\paragraph{Lossless structural merge.}
For a confirmed merge \(a\rightarrow b\), the system unions the episodic and
semantic row identifiers, keeps aliases, sums mention counts, keeps the maximum
admission score, and rewires every role-typed frame arm to \(b\). Union, sum, and maximum are commutative and
associative, so the order of repeated merges does not change the reachable
evidence. The dropped anchor is removed only after its values and edges have
been transferred.

Semantic Dream never rewrites the memories in the vector database, and it consolidates neither
episodic-only anchors nor fact nodes. The asymmetry is deliberate. Semantic
identity can be refined incrementally, whereas event occurrences and their
citations are kept as observations. AutoDream therefore makes the graph less
fragmented without creating a textual summary whose embedding could hide the
details of its inputs.

\paragraph{Procedural refinement.}
For procedural memory, AutoDream refines instead of merging. A distiller judges
the outcome of each pending session from quotes in its transcript and distills
experiences \(e_1,\ldots,e_m\), each marked as a lesson worth reusing or a
mistake worth avoiding. A curator~\citep{zhang2026ace} turns these experiences
into skills in the procedural store \(P_u^P\), and it can create, edit, or
delete a skill. An edit can cite the experiences behind it with a quote of the
text it adds, and a citation is recorded only when that quote checks out, so a
cited step of a skill can be traced through its experiences to the quoted
sessions they came from. In \Cref{fig:mirix-agent-centric}, one failed purchase
yields an experience to avoid and one to reuse, and the curator adds two steps
to the shopping routine that cite the first and keeps a step that the second
supports. The distiller's unclear verdict and the curator's non-edit verdicts are in \Cref{app:method-autodream}.

\FloatBarrier
\section{Experiments}
\label{sec:experiments}

We evaluate \ours for the three properties it is designed for. Accuracy is
measured for user memory on conversational recall, document QA, and
personalized generation (\Cref{sec:exp-user}), which mainly exercise semantic
memory together with the episodic records that ground it, and for agent memory
on held-out interactive tasks and on
MetaClaw-Bench, whose tasks span 30 days (\Cref{sec:exp-agent}), which exercise procedural memory. Efficiency is measured
as ingestion time, answer latency, and answer tokens for user memory, with
AutoDream disabled in every timed run, and as training and test tokens for
agents (\Cref{sec:exp-efficiency}). Practicality is built into the protocol itself. One system serves both uses, no model is fine-tuned, the agent
evaluation covers closed and open backbones, and every graph configuration runs
under the budgets of \Cref{sec:retrieval}.

\subsection{Experimental setup and system variants}
\label{sec:exp-setup}

\paragraph{Default system and ablations.}
Unless a configuration is explicitly identified otherwise, \ours denotes the
full architecture with the provenance hypergraph and AutoDream enabled.
For the user-memory ablations, \textbf{M0} uses flat vector memory without the
graph or AutoDream, like MIRIX-1.0~\citep{mirix}, and the tables label it
MIRIX-1.0. \textbf{M1} adds graph indexing and retrieval but disables
AutoDream, and \textbf{M2} is the full graph plus AutoDream configuration.
M1 and M2 are configurations of \ours, not successive releases of the
method. In M2, semantic AutoDream is invoked by the ingestion driver after
every five chunks and after the final chunk, outside the critical write path (Section~\ref{sec:autodream} and \Cref{app:method-autodream}). It merges eligible entity anchors while preserving
stored records and episodic occurrences.

The conversational tables report all three configurations. The available
document QA results and graph efficiency measurements use M1 and are labeled
accordingly; disabling AutoDream there is an experimental choice, not a
restriction that prevents consolidation of document memories. MemoryCD is
reported under the system name, but its recorded results do not identify an
M0, M1, or M2 configuration, so we do not use them to estimate the contribution
of AutoDream. For agent memory, we evaluate the procedural refinement path,
in which AutoDream distills sessions into reusable skills whose retrieval is
independent of the semantic hypergraph. The M0, M1, and M2 comparisons below
therefore concern user-memory indexing and semantic consolidation, rather
than a procedural graph ablation.

\paragraph{User-memory tasks and baselines.}
LoCoMo~\citep{locomo} and LongMemEval-S~\citep{longmemeval} test conversational
recall. SHDocQA, MHDocQA, and DetectiveQA in the MemoryAgentBench
evaluation~\citep{memoryagentbench} extend this analysis to factual and
relational evidence in documents. They are evidence-retrieval tests rather
than direct personalization tasks. MemoryCD~\citep{memorycd} tests whether
stored user information supports personalized review summarization and
generation. We compare with mem0, A-Mem, MemoryBank, Graphiti, and HyperMem on
QA, and with the methods listed in the MemoryCD tables, including
long-context and RAG baselines, on personalization.

\paragraph{Models and scoring.}
For the user-memory evaluation, the answerer and extractor use
\texttt{gpt-4.1-mini} with temperature 0 and seed 42, and the topic extractor
uses \texttt{gpt-4.1-nano}. \ours uses \texttt{text-embedding-3-small}; external
baselines retain their default embeddings. SHDocQA and MHDocQA use substring
matching, and DetectiveQA uses letter exact matching. LoCoMo and LongMemEval-S
use a \texttt{gpt-4o-mini} judge with majority-of-3 voting. MemoryCD uses
ROUGE-1, ROUGE-L, BLEU-4, and BERTScore-F1. Agent backbones and evaluation
protocols are specified in Section~\ref{sec:exp-agent}. Differences in accuracy
or success rate are reported in percentage points.

\paragraph{Efficiency and uncertainty.}
We report total ingestion wall-clock time, mean time to retrieve and answer a
question, total answer tokens, and graph node and edge counts where available.
Graph size measures index structure, not stored text length or total storage.
For interactive agents, we report success, action steps, and training and test
tokens, with mean and standard deviation over three runs. For LoCoMo, all
comparisons below use the results in Table~\ref{tab:locomo}; M0, M1, and M2
are distinct system configurations, not repeated runs used to estimate
variance.

\subsection{User memory: recall, documents, and personalization}
\label{sec:exp-user}

% [24] = narrow lines, one past the caption so the text below it has room; change it if the caption gets longer or shorter
\begin{wrapfigure}[24]{r}{0.49\textwidth}
\vspace{-1.1\baselineskip}
\centering
\includegraphics[width=\linewidth]{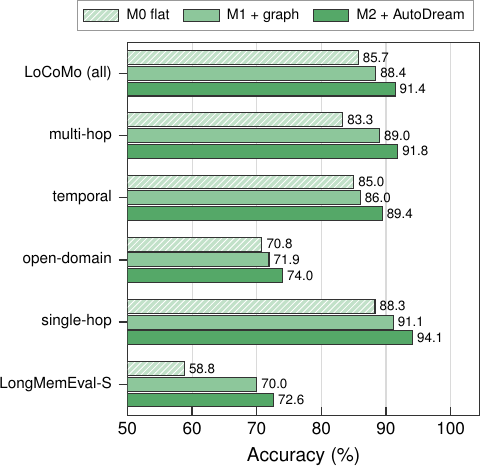}
\caption{Components of \ours on LoCoMo, its four question types, and LongMemEval-S. The hypergraph (M1) and AutoDream (M2) each add accuracy over flat memory (M0).}
\label{fig:res-ablation}
\vspace{0.3\baselineskip}
\end{wrapfigure}
\paragraph{The hypergraph and AutoDream each improve conversational recall.}
\Cref{fig:res-ablation} and \Cref{tab:locomo}, with \Cref{tab:lme} in the appendix, compare the M0,
M1, and M2 configurations under the evaluation protocol described above and show a
consistent ordering on both conversational benchmarks. On LoCoMo, graph retrieval raises overall accuracy
from 85.7\% for M0 to 88.4\% for M1, and semantic AutoDream raises it to
91.4\% for M2. On LongMemEval-S, the corresponding values are 58.8\%, 70.0\%,
and 72.6\%. The graph increment is thus 2.7 points on LoCoMo and 11.2 points
on LongMemEval-S, while AutoDream adds 3.0 and 2.6 points, respectively.
The larger graph increment on LongMemEval-S suggests that evidence organization
can matter more when flat retrieval has difficulty recovering the relevant
history. It does not establish which question types account for that gain,
since a LongMemEval-S category breakdown is not available here.

\begin{table*}[t]
\centering
\caption{LoCoMo accuracy (\%) by question type on 1{,}540 non-adversarial questions. Bold marks the best value in each column.}
\label{tab:locomo}
\small
\begin{tabular}{lccccc}
\toprule
Method & Overall & Multi-hop & Temporal & Open & Single-hop \\
\midrule
mem0        & 82.9 & 80.5 & 80.4 & 66.7 & 86.6 \\
A-Mem       & 68.3 & 68.8 & 55.5 & 61.5 & 73.8 \\
MemoryBank  & 59.5 & 70.9 & 14.6 & 52.1 & 73.6 \\
Graphiti    & 58.4 & 65.2 & 37.1 & 53.1 & 64.9 \\
HyperMem    & 91.0 & 91.1 & 84.4 & \textbf{76.0} & \textbf{95.2} \\
\midrule
MIRIX-1.0 (M0)                & 85.7 & 83.3 & 85.0 & 70.8 & 88.3 \\
\ours\ (M1, graph)             & 88.4 & 89.0 & 86.0 & 71.9 & 91.1 \\
\ours\ (M2, graph $+$ dream)   & \textbf{91.4} & \textbf{91.8} & \textbf{89.4} & 74.0 & 94.1 \\
\bottomrule
\end{tabular}
\end{table*}
\paragraph{Semantic consolidation can help temporal recall without rewriting events.}
The LoCoMo breakdown is more informative than overall accuracy alone. Moving
from M0 to M1 improves multi-hop accuracy by 5.7 points, compared with 1.0
point on temporal questions. Moving from M1 to M2 then improves every
reported category, including 3.4 points on temporal questions. This pattern is
consistent with the division of work in the method. N-ary frames connect
related evidence, while identity consolidation gives retrieval a more coherent
entry point into records that mention the same entity. Since semantic
AutoDream preserves fact nodes and episodic occurrences, temporal evidence can
become easier to reach without collapsing distinct events. This is an
interpretation of the aggregate results; a direct audit of retrieved records
and merged anchors would be needed to establish the mechanism for individual
answers.

\vspace{4.5pt}% room between the full-width line above and Figure 5
% [17] = narrow lines, one past the caption so the text below it has room; change it if the caption gets longer or shorter
\begin{wrapfigure}[17]{l}{0.49\textwidth}
\vspace{-1.1\baselineskip}
\centering
\includegraphics[width=\linewidth]{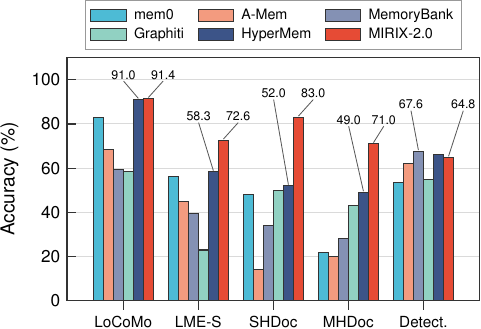}
\caption{\ours against memory baselines on five QA benchmarks. The document tasks use M1.}
\label{fig:res-baselines}
\vspace{-0.6\baselineskip}
\end{wrapfigure}
\Cref{fig:res-baselines} compares \ours with all memory baselines on the five QA benchmarks. M2
exceeds HyperMem by 0.4 points overall on LoCoMo and by 14.3 points on
LongMemEval-S. Its LoCoMo advantage is concentrated in temporal and multi-hop
questions, with gains of 5.0 and 0.7 points, while HyperMem retains higher
open-domain and single-hop scores. The evidence therefore supports stronger
recovery of connected and time-grounded facts, rather than dominance on every
form of recall. HyperMem's 91.0\% comes from our own run and is lower than the
92.7\% it reports. \Cref{app:results-user} describes how we ran HyperMem with
its original settings.

\paragraph{Keeping source details improves document QA.}
\Cref{tab:docqa} in the appendix reports M1 on document QA. Its 83.0\% on SHDocQA and
71.0\% on MHDocQA exceed HyperMem by 31.0 and 22.0 points. Gains on both tasks
are consistent with a design that retains source details as well as relational structure. Single-hop questions require the precise fact, while multi-hop
questions also require connections between supporting records. In
Sections~\ref{sec:typed-evidence} and~\ref{sec:retrieval}, these requirements
motivate source-faithful writers and retrieval of original rows alongside
graph facts. The comparison measures the combined pipeline and does not
isolate either feature.

DetectiveQA exposes a boundary of this advantage. M1 reaches 64.8\%, below
MemoryBank's 67.6\% and HyperMem's 66.2\%. Organizing explicit evidence does
not by itself guarantee the inference needed to resolve a detective question.
Without an error analysis, we cannot distinguish missed evidence from failures
to reason over retrieved evidence. The document results also cannot establish
an AutoDream benefit, because they evaluate M1 only.
\paragraph{Selective user memory supports personalized generation.}
\Cref{tab:memcd-summ}, and \Cref{tab:memcd-gen} in the appendix, move from answering factual
questions to generating text for a particular user. Cross-domain evaluation is especially relevant because useful preferences must transfer from other product
domains to Home\_and\_Kitchen. This tests a different role for semantic memory
from recalling a named event. A useful record must preserve a user's relevant
preferences while remaining accessible when the new request has a different
surface topic.

For cross-domain summarization, \ours has the highest value on all four
metrics. Its ROUGE-L is .1510, compared with .1430 for the strongest baseline
on that metric, and its BLEU-4 is .0209, compared with .0162. In the
single-domain setting it leads on ROUGE-1 and ROUGE-L and ties the highest
BERTScore, while long-context input has higher BLEU-4. Review generation
shows the highest reported scores for \ours in both settings on all four
metrics, although several margins are small. Together, these results suggest
that selective user memory can preserve useful information for generation
without requiring the full history as context. They do not measure factual
consistency, preference satisfaction, or user approval directly, and the
available configuration record does not support a MemoryCD-specific claim
about graph or AutoDream gains.

\begin{table*}[t]
\centering
\caption{Review summarization on MemoryCD (100 Home\_and\_Kitchen users). Bold marks column maxima.}
\label{tab:memcd-summ}
\small
\begin{tabular}{lcccc|cccc}
\toprule
& \multicolumn{4}{c|}{Cross-domain} & \multicolumn{4}{c}{Single-domain} \\
\cmidrule(lr){2-5}\cmidrule(lr){6-9}
Method & R-1 & R-L & B-4 & BERT & R-1 & R-L & B-4 & BERT \\
\midrule
Long-context & .1493 & .1357 & .0162 & .8745 & .1675 & .1570 & \textbf{.0249} & \textbf{.8779} \\
RAG          & .1559 & .1430 & .0145 & .8735 & .1670 & .1523 & .0196 & .8759 \\
mem0         & .1526 & .1420 & .0159 & .8749 & .1593 & .1472 & .0184 & .8767 \\
A-Mem        & .1567 & .1428 & .0160 & .8746 & .1602 & .1521 & .0184 & .8764 \\
MemoryBank   & .1409 & .1291 & .0131 & .8736 & .1584 & .1449 & .0188 & .8760 \\
Graphiti     & .1534 & .1415 & .0139 & .8741 & .1520 & .1418 & .0172 & .8746 \\
\ours\  & \textbf{.1603} & \textbf{.1510} & \textbf{.0209} & \textbf{.8754} & \textbf{.1715} & \textbf{.1617} & .0234 & \textbf{.8779} \\
\bottomrule
\end{tabular}
\end{table*}

\subsection{Agent memory: held-out tasks and skills}
\label{sec:exp-agent}

\paragraph{Setup and transfer to held-out tasks.}
We evaluate procedural memory on ALFWorld~\citep{shridhar2021alfworld} and
WebShop~\citep{yao2022webshop}, comparing an agent without memory with
ReasoningBank~\citep{ouyang2026reasoningbank}, Agent Workflow Memory
(AWM)~\citep{wang2025awm}, and \ours. Within each comparison, methods share the
actor prompt and backbone. ALFWorld uses GPT-6 Luna and GPT-5.6 Luna; WebShop
uses GPT-6 Luna. Each memory method learns from the same training tasks,
39 for ALFWorld and 40 for WebShop. Memory is then frozen for evaluation on
134 held-out ALFWorld episodes with a 50-step cap and 140 WebShop instructions
with a 15-step cap. Results are averaged over three independent runs.
The separation between memory learning and evaluation tests reuse of acquired
procedures on new tasks.

This setting exercises the procedural branch of AutoDream
(Section~\ref{sec:autodream}). Sessions supply lessons to reuse and mistakes to
avoid, and a curator turns those experiences into skills. Unlike semantic
consolidation, which preserves the factual records and changes their index,
procedural refinement changes the instructions available to the actor. The
experiments evaluate the utility of that combined process; they do not
separately ablate the outcome judge, experience distiller, or curator.

\begin{wrapfigure}{r}{0.49\textwidth}
\vspace{-1.1\baselineskip}
\centering
\includegraphics[width=\linewidth]{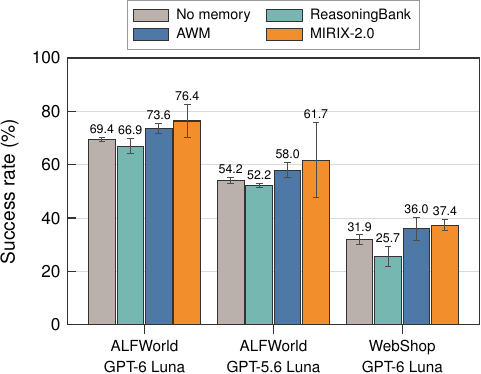}
\caption{Agent memory on held-out tasks. Success rate, mean and standard deviation over three runs.}
\label{fig:res-agent}
\vspace{-0.6\baselineskip}
\end{wrapfigure}
\begin{wrapfigure}[18]{L}{0.49\textwidth}
\vspace{-1.1\baselineskip}
\centering
\includegraphics[width=\linewidth]{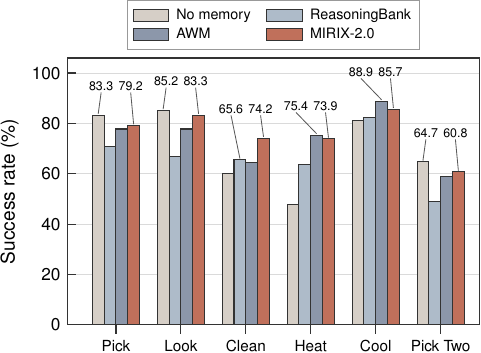}
\caption{ALFWorld success by task type with GPT-6 Luna, over three runs.}
\label{fig:res-alfworld-types}
\vspace{-0.6\baselineskip}
\end{wrapfigure}

\paragraph{ALFWorld: useful procedures must generalize across action sequences.}
\Cref{fig:res-agent}, and \Cref{tab:alfworld-proc} in the appendix, show the highest mean
overall success among the compared methods for \ours with both backbones. With GPT-6 Luna it reaches
76.4\%, compared with 69.4\% without memory and 73.6\% with AWM. With GPT-5.6
Luna it reaches 61.7\%, compared with 54.2\% and 58.0\%, respectively.
The gains over no memory are similar, at 7.0 and 7.5 points, but they are not
uniform across task types (\Cref{fig:res-alfworld-types}). With GPT-6 Luna, Clean improves from 60.2\% to
74.2\% and Heat from 47.8\% to 73.9\%, while Pick, Look, and Pick Two remain
below the no-memory scores. This suggests that procedures are most useful
when they supply missing action structure; retrieving a routine can also
interfere with a task the actor already handles well.

The step counts provide a complementary view. Successful episodes require
13.2 rather than 14.2 steps with GPT-6 Luna and 12.4 rather than 16.1 with
GPT-5.6 Luna, relative to no memory. This is consistent with more direct
action sequences, but the successful subsets differ across methods, so these
are not paired measurements on identical solved tasks. The large success-rate
standard deviations for \ours, 6.2 and 14.0 points, also limit the conclusion. It has the highest observed mean, but three runs do not establish a stable
advantage over AWM.

\paragraph{WebShop: procedure quality matters more than memory availability.}
\Cref{tab:webshop-proc} in the appendix reports 37.4\% success for \ours, compared with
31.9\% without memory, 25.7\% with ReasoningBank, and 36.0\% with AWM.
ReasoningBank falls below no memory on both interactive benchmarks with
GPT-6 Luna, showing that adding experience memory is not sufficient to improve
an actor. In WebShop, \ours leads the memory methods on Fashion and Beauty,
ties AWM on Garden and Grocery, and trails AWM on Electronics. A plausible
interpretation is that reusable constraint checks help some product searches,
while other searches need more task-specific guidance. The category results
do not directly identify which learned skills caused those differences.

\ours uses 8.2 steps per episode and 6.3 per successful episode, fewer than
the other memory methods but slightly more than no memory at 7.9 and 6.2.
Thus, its benefit is a higher success rate with modest additional interaction,
not a reduction in steps relative to every baseline.

\vspace{8pt}% room between the full-width line above and Figure 7
% [18] = narrow lines, one past the caption so the text below it has room; change it if the caption gets longer or shorter
\paragraph{MetaClaw-Bench: answer quality and completed artifacts are distinct.}
\Cref{tab:metaclaw_part1} evaluates agent memory on Part~I of
MetaClaw-Bench~\citep{xia2026metaclaw}, which spans 30 days.
Accuracy is the official mean
per-question score, which combines binary file checks with partial credit on
multiple-choice questions, and completion is the completion rate over the 224 file-check
questions. With GPT-5.2, \ours improves the official
mean question score from 44.0\% for MetaClaw Skills to 49.2\%, and file-check
completion from 17.1\% to 27.7\%. The larger completion gain, 10.6 points,
suggests that reusable procedures can help produce the required artifacts as
well as improve question scores. With Qwen3.6-35B-A3B, the score gain is
smaller, from 51.8\% to 52.7\%, and completion is tied at 30.8\%. The benefit
therefore depends on the backbone and on what the metric measures.
The GPT-5.2 baselines are reported values from the MetaClaw paper, so this
comparison is not evidence from a fully controlled rerun. Unlike ALFWorld and
WebShop, this table does not report variation across independent runs or a
matched procedural-component ablation.

\begin{table}[tb]
\centering
\captionsetup{justification=centering}
\caption{Agent memory on MetaClaw-Bench Part~I (346 questions).}
\label{tab:metaclaw_part1}
\small
\begin{tabular}{llcc}
\toprule
Model & Skill module & Acc.\ (\%) & Compl.\ (\%) \\
\midrule
\multirow{3}{*}{GPT-5.2}
  & Baseline        & 41.1 & 14.7 \\
  & MetaClaw Skills & 44.0 & 17.1 \\
  & \ours                 & \textbf{49.2} & \textbf{27.7} \\
\midrule
\multirow{3}{*}{Qwen3.6-35B-A3B}
  & Baseline                  & 40.3 & 12.1 \\
  & MetaClaw Skills           & 51.8 & \textbf{30.8} \\
  & \ours                 & \textbf{52.7} & \textbf{30.8} \\
\bottomrule
\end{tabular}
\end{table}

\subsection{Efficiency: ingestion time, answer latency, and tokens}
\label{sec:exp-efficiency}

\paragraph{Graph construction moves work into ingestion.}
\Cref{fig:res-cost}a and \Cref{tab:eff-docqa} in the appendix show the cost of organizing
source records before queries arrive. Relative to M0, M1 increases ingestion time by
3.8 to 4.0 times across the three document tasks. M1 also has higher answer
latency than M0 on each task, so graph retrieval is not intrinsically cheaper
than flat retrieval. Its purpose is to improve access to structured evidence,
with extraction and indexing paid in advance and additional evidence handling
paid at query time.

Against HyperMem, M1 answers faster on all three document tasks
(\Cref{fig:res-cost}b), at 5.45
versus 10.55 seconds on SHDocQA, 9.01 versus 10.65 on MHDocQA, and 7.24 versus
13.05 on DetectiveQA. It uses fewer answer tokens on SHDocQA and MHDocQA,
but more on DetectiveQA, at 1.16M versus 0.35M. The DetectiveQA result shows that lower latency can coexist with higher token use
and lower accuracy. Latency, token cost, and answer quality therefore need to
be assessed separately.

\begin{figure}[t]
\centering
\includegraphics[width=\linewidth]{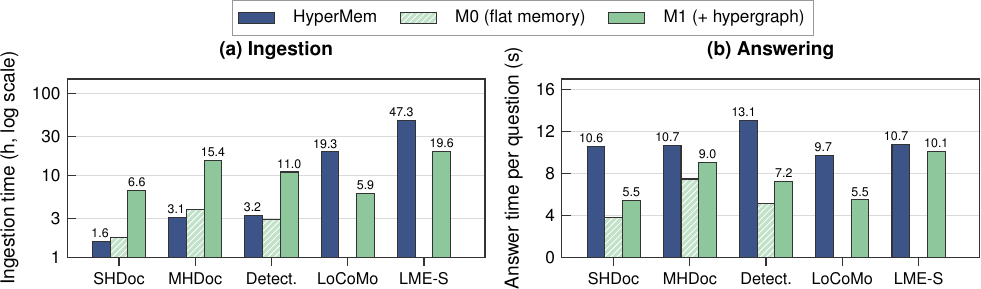}
\caption{Cost of user memory: (a) ingestion time and (b) answer time per question.}
\label{fig:res-cost}
\end{figure}
\paragraph{The available timings characterize M1, not the full M2 system.}
\Cref{fig:res-cost}b and \Cref{tab:eff-docqa} in the appendix show that M1 also has lower
answer latency than HyperMem on LoCoMo and \mbox{LongMemEval-S}, completing the five-benchmark latency
comparison. Its conversational ingestion time is lower as well, unlike the
document setting. These measurements are consistent with bounded graph access keeping answer
latency low, but they do not isolate the effects of candidate caps, model
calls, or execution conditions. M2 adds consolidation calls and can
change subsequent retrieval behavior. Its total ingestion cost, answer cost,
and graph reduction must be measured directly before attaching the M1
latency claims to the full system. Node and edge counts alone likewise do
not quantify evidence retention or compression.
\paragraph{Fewer agent steps do not necessarily mean fewer tokens.}
\Cref{tab:agent-token-consumption} separates the cost of learning a procedure from the
cost of applying it, with the token accounting given in \Cref{app:results-cost}. On WebShop, \ours uses the fewest
training and test tokens among the memory methods, 1.815M and 4.975M per run,
while achieving the highest mean success. On ALFWorld, its 3.813M test tokens
are the highest among the memory methods despite its lower mean step count.
A retrieved procedure can reduce interaction steps while increasing the
context processed at each step. Relative to no memory, test tokens rise on
both benchmarks, so procedural memory should be judged by the resulting
success and reuse, rather than assumed to save tokens automatically.
The reported totals do not determine how many future tasks are needed to
amortize memory learning, because no varying-horizon cost study is included.

\begin{table*}[tb]
\centering
\caption{Token use per run and success with GPT-6 Luna, mean $\pm$ std over three runs. Bold marks the best of the three memory methods.}
\label{tab:agent-token-consumption}
\small
\setlength{\tabcolsep}{7pt}
\begin{tabular}{llccc}
\toprule
Benchmark & Method
& Training tokens (M) $\downarrow$
& Test tokens (M) $\downarrow$
& Success rate (\%) $\uparrow$ \\
\midrule
\multirow{4}{*}{ALFWorld}
& No memory
& 0
& 2.147$\pm$0.048
& 69.4$\pm$0.7 \\
& ReasoningBank
& \textbf{1.747$\pm$0.122}
& \textbf{3.351$\pm$0.045}
& 66.9$\pm$2.8 \\
& AWM
& 2.749$\pm$0.059
& 3.674$\pm$0.332
& 73.6$\pm$1.9 \\
& \ours
& 2.253$\pm$0.100
& 3.813$\pm$0.490
& \textbf{76.4$\pm$6.2} \\
\midrule
\multirow{4}{*}{WebShop}
& No memory
& 0
& 4.238$\pm$0.463
& 31.9$\pm$1.8 \\
& ReasoningBank
& 2.010$\pm$0.242
& 6.620$\pm$0.674
& 25.7$\pm$3.8 \\
& AWM
& 1.909$\pm$0.127
& 5.448$\pm$0.356
& 36.0$\pm$4.3 \\
& \ours
& \textbf{1.815$\pm$0.054}
& \textbf{4.975$\pm$0.340}
& \textbf{37.4$\pm$2.1} \\
\bottomrule
\end{tabular}
\end{table*}

\subsection{What the experiments establish}
\label{sec:exp-discussion}

The results support the three properties in different measure. Accuracy has
the broadest support. The M0, M1, and M2 ordering holds on both conversational
benchmarks, and \ours leads the compared systems on four of the five QA
benchmarks, two of them with AutoDream disabled. As procedural memory, it has
the highest mean success among the compared memory methods on ALFWorld with
both backbones and on WebShop, and the highest MetaClaw-Bench score with both
backbones, while the no-memory comparisons show that retrieved experience can
also hurt. Efficiency holds for answering more than for ingestion. With
AutoDream disabled, \ours answers faster than HyperMem on every timed
benchmark, and on WebShop it uses the fewest training and test tokens among
the memory methods. Graph construction, however, makes document ingestion 3.8
to 4.0 times slower than flat memory, and the ALFWorld test tokens of \ours are
the highest among the memory methods. Practicality rests on the design and on
our protocol rather than on a deployment. One store serves both uses, no model
is fine-tuned, and every graph fact returned to the answerer cites its source
rows.

The tables evaluate complete configurations, so the component explanations in
this section remain interpretations, and the limitations in
\Cref{sec:conclusion} list the ablations that would test them. The tables also
do not measure merge precision, source-detail retention, skill provenance
coverage, or the contribution of the six-writer routing policy, and they do not
report memory length or user-memory ingestion tokens, for which graph counts
and answer tokens cannot substitute.

\DeferredFloatBarrier
\section{Related Work}
\label{sec:related}

\paragraph{Memory systems for LLM agents.}
Memory systems for LLM agents keep past interactions outside the model and
retrieve them when they become
relevant~\citep{agentmemorysurvey,zhang2024memorysurvey,lscs,guo2026multimodalmemory}.
Early designs rank or decay a memory stream~\citep{park2023generative,MemoryBank}
or manage memory the way an operating system manages
storage~\citep{memgpt,memoryos,li2025memos}. Later systems extract and update
salient facts~\citep{mem0}, link notes into an evolving network~\citep{amem},
compress what they store~\citep{fang2026lightmem,liu2026simplemem}, or learn
memory construction with reinforcement
learning~\citep{wang2025memalpha,yan2026memoryr1}. Graph memories organize
content around entities and their
relations~\citep{bei2025graphsagents,zep,gutierrez2024hipporag,gutierrez2025hipporag2,edge2024graphrag,wu2026gam},
and hypergraph memories keep n-ary facts
together~\citep{luo2025hypergraphrag,hypermem}. Typed designs split memory by
function, following cognitive accounts of episodic and semantic
memory~\citep{tulving1985memory,episodic_memory_missing_piece}.
MIRIX~\citep{mirix} assigns six memory types to dedicated agents, and
PersonaAgent~\citep{zhang2026personaagent} pairs episodic and semantic memory
for personalization. Memory stored in model parameters or latent
states~\citep{memoryllm,mplus,self-param,behrouz2024titans} requires training
and cannot be used with closed models. Evaluation has moved from recall over
long conversations~\citep{locomo,longmemeval} to multi-session agentic
tasks~\citep{memoryagentbench,he2026memoryarena}, evolving
environments~\citep{ji2026clawarena}, lifelong
personalization~\citep{memorycd}, memory audits~\citep{ma2026memaudit}, and
comparisons of memory substrates~\citep{harnessmemory}. \ours combines the
typed and the graph directions. Complete typed records remain the system of
record, a provenance hypergraph serves only as their index, and every graph
fact cites its source records.

\paragraph{Multi-agent systems.}
Multi-agent systems divide a task among LLM agents that take different
roles~\citep{guo2024masurvey,luo2025agentsurvey}, from role-playing and
conversable agents~\citep{li2023camel,wu2024autogen} to software teams and
groups of experts~\citep{hong2024metagpt,qian2024chatdev,chen2024agentverse}.
Agents can debate~\citep{du2024debate,liang2024debate,huang2026madiave},
optimize their workflow~\citep{zhuge2024gptswarm,zhang2025aflow}, or
communicate through latent representations~\citep{yang2026xbridge}. They also
fail in new ways~\citep{cemri2025mas,huang2026believed}, and one response keeps
humans in the loop as
collaborators~\citep{zou2026hasurvey,zou2025collabintel,wu2026hasbench}. Memory
in these systems mostly serves the collaboration, through shared pools, graphs
of team trajectories, and sharing across users under access
control~\citep{memory_sharing,zhang2025gmemory,rezazadeh2025collabmemory}.
\ours applies the multi-agent design to memory itself. A Meta Memory Manager
routes each input to the agents of the six memory types, and each agent writes
only the records of its own type.

\paragraph{Self-reflection and memory consolidation.}
Self-reflection lets an agent improve from feedback on its own behavior without
updating its
weights~\citep{shinn2023reflexion,madaan2023selfrefine,park2023generative,zhao2025marco}.
Experience-driven methods turn such feedback into reusable knowledge, including
insights~\citep{zhao2024expel}, executable skills~\citep{wang2024voyager},
workflows~\citep{wang2025awm}, reasoning
strategies~\citep{ouyang2026reasoningbank}, and evolving playbooks and case
banks~\citep{suzgun2025cheatsheet,zhang2026ace,fang2026memp,zhou2025memento}.
Skills can co-evolve with the policy~\citep{xia2026skillrl,xia2026metaclaw}, be
verified and optimized as
artifacts~\citep{zhang2026coevoskills,ma2026skillneo}, or form a
hypergraph~\citep{xu2026hyperskill}, and agents, their harnesses, and their
memory architectures can improve
themselves~\citep{xia2026agent0,xia2026rrsi,liu2026evolvemem,gao2026selfevolving,fang2025selfevolving}.
Consolidation can also run offline, before queries arrive or while the user is
inactive~\citep{lin2025sleeptime,fang2026lightmem,xia2026metaclaw,ye2026autodreamer}.
Rewriting memory carries risks. Consolidated memories can become
faulty~\citep{zhang2026faultymemories}, and a defective skill can contaminate
the skills distilled after it~\citep{shang2026backfires}. AutoDream also
consolidates outside the write path, but it limits what consolidation may
change. On semantic memory it merges only anchors that denote the same entity
and never rewrites stored records. On procedural memory it keeps a judged
outcome, and each experience a skill edit cites, only after checking the
quoted text.

\section{Conclusion}
\label{sec:conclusion}

We presented \ours, a multi-agent memory system that keeps complete typed
records, indexes them with a provenance hypergraph under explicit budgets, and
runs AutoDream outside the write path to merge duplicate anchors and refine
skills. It leads the compared systems on four of the five QA benchmarks, two of
them with AutoDream disabled, reaching 91.4\% on LoCoMo and 72.6\% on
LongMemEval-S, and it has the highest value on all four metrics for
cross-domain review summarization on MemoryCD. On LoCoMo, the hypergraph and
AutoDream each add accuracy over flat memory. As procedural memory, its
distilled skills lead the compared memory methods in mean success on ALFWorld
and WebShop, and with GPT-5.2 they reach 49.2\% on MetaClaw-Bench, against
44.0\% reported for the MetaClaw skill library. With AutoDream disabled, \ours
answers each question in less time than HyperMem on all five timed benchmarks,
and on WebShop its skills cost the fewest tokens among the memory methods. One
store serves both uses, and the system runs with closed and open models without
fine-tuning. The results suggest that a memory layer can be
practical, accurate, and efficient at once when complete records remain the
system of record, the graph is only a bounded index, and consolidation changes
how evidence is connected rather than what evidence exists. 

\paragraph{Limitations.}
Our ablations add components in groups. They do not isolate n-ary frames,
cited source rows, occurrence identities, answer-time verification, or the
judge, distiller, and curator of procedural memory, so the explanations in \Cref{sec:experiments}
remain interpretations. The user-memory results come from single runs, partly
scored by an LLM judge, the agent results average three runs with large
spreads, and some baselines are values reported by their authors. 

% Building the
% graph slows the ingestion of long documents, the cost of AutoDream is not yet
% measured, and the public benchmarks we use do not test months of real use.

\section*{Ethics and Broader Impact Statement}

\ours lets assistants and task agents carry what users told them, and what the
agents learned, from one session to the next. Every graph fact cites the records it was
extracted from, so an answer can be traced to stored evidence.
It is intended for assistants that users knowingly choose to use and should
not be used to build profiles of people without their consent. A user-centric
memory accumulates personal and possibly sensitive information. \ours keeps
sensitive verbatim values in a Knowledge Vault that the hypergraph does not
index, but this does not replace consent and access control. A deployment
should let users view, correct, and delete their memories, confine each memory
to its user~\citep{rezazadeh2025collabmemory}, and use model providers or local
models that meet its privacy requirements, because stored content is sent to
the model. Memory can also make mistakes
persistent~\citep{zhang2026faultymemories,shang2026backfires}, and content
injected into a conversation can be stored and later retrieved as if it were
trusted. A wrong merge
changes only the index, and a procedural edit keeps a citation only after its
quote is checked, but in high-stakes settings learned skills and merged
identities should still be reviewed by a person and stored memories audited
periodically~\citep{ma2026memaudit}. All experiments use publicly released
benchmarks under their licenses. We collected no new data and ran no study with human participants.
MemoryCD is built from public Amazon product reviews, and we use it as
released without attempting to identify the reviewers.

% \section*{Acknowledgments}
\newpage
\bibliographystyle{plainnat}
\bibliography{ref}

%%%%%%%%%%%%%%%%%%%%%%%%%%%%%%%%%%%%%%%%%%%%%%%%%%%%%%%%%%%%%%%%%%%%%%%%%%%%%%%
%%%%%%%%%%%%%%%%%%%%%%%%%%%%%%%%%%%%%%%%%%%%%%%%%%%%%%%%%%%%%%%%%%%%%%%%%%%%%%%
% APPENDIX
%%%%%%%%%%%%%%%%%%%%%%%%%%%%%%%%%%%%%%%%%%%%%%%%%%%%%%%%%%%%%%%%%%%%%%%%%%%%%%%
%%%%%%%%%%%%%%%%%%%%%%%%%%%%%%%%%%%%%%%%%%%%%%%%%%%%%%%%%%%%%%%%%%%%%%%%%%%%%%%

\appendix
\numberwithin{equation}{section}
% A page that holds only floats starts them at the top instead of centering them.
\makeatletter\setlength{\@fptop}{0pt}\makeatother

\newpage

\section{Method details}
\label{app:method}

This appendix collects the details of \Cref{sec:method} that the main text summarizes or leaves out, in the order of its
subsections. For each part it states the
constants, schedules, scoring rules, and examples, explains why they are
designed as they are, and names the step of \Cref{alg:mirix} they belong to.

\subsection{Memory types, routing, and writing}
\label{app:method-routing}

The Meta Memory Manager is the first call of the write path
(\Cref{alg:mirix}, line~\ref{alg:route}). The router reads the input and
selects the memory types it updates. Each selected writer then reads the
current store \(P_u^{t-1}\) and produces a record that line~\ref{alg:commit}
commits to the evidence plane before any graph work begins. Formally, each
input \(x_t\) is mapped to a subset of six specialized writers,
\begin{equation}
    \begin{aligned}
    \mathcal K_t&=R_\theta(x_t),\\
    m_t^{(k)}&=W_k(x_t,P_u^{t-1}),\quad k\in\mathcal K_t,
    \end{aligned}
    \label{eq:routing}
\end{equation}
and each writer owns the schema and update operations of one memory type. Both
steps of \eqref{eq:routing} are language-model calls, which is why the router
and the writers carry the \(\star\) mark in \Cref{alg:mirix}.
\Cref{tab:memory-types} lists the six types.

\suppressfloats[t]
\begin{table}[t]
\centering
\caption{The six memory types of \ours and their roles.}
\label{tab:memory-types}
\small
\begin{tabular}{lll}
\toprule
Memory & Primary role & Hypergraph indexed \\
\midrule
Core & persistent user and agent state & no \\
Episodic & time-grounded occurrences & yes \\
Semantic & durable facts and concepts & yes \\
Procedural & executable workflows & no \\
Resource & documents and multimodal artifacts & no \\
Knowledge Vault & sensitive verbatim values & no \\
\bottomrule
\end{tabular}
\end{table}

The separation is functional. Only episodic and semantic memory are indexed by
the hypergraph, because they hold the event and relational evidence that
cross-memory traversal needs. The other types keep their direct, type-specific
retrieval paths. Core memory holds the persistent state of the user and the
agent, procedural memory holds the skills that \Cref{sec:autodream} refines,
resource memory holds documents and multimodal artifacts, and the knowledge
vault holds sensitive verbatim values. Each of them is read through its own typed path rather than through the graph.
In \Cref{alg:mirix}, the test \(k\in\{\mathrm{episodic},\mathrm{semantic}\}\) is the only step of the write
path that depends on whether a record is graph-indexed. In \Cref{fig:mirix-user-centric}, both user turns are
routed to the episodic and semantic writers.

\paragraph{Source-faithful writing.}
Before a record reaches the graph extractor, the writer prompts of our runs ask
the episodic and semantic writers to keep complete lists, names, counts, titles,
locations, and dates in the record's details instead of replacing them with a
topical gist. Episodic writers store the time an event happened rather than the
time it was mentioned, and they store a planned event with its planned date.

\paragraph{Write ordering.}
The write to the vector database commits before graph construction begins
(line~\ref{alg:commit}). Graph extraction
is a fail-soft indexing hook, so a graph failure is logged but never rolls back
a stored memory. The evidence plane is thus the durable system of record rather
than a side effect of graph construction.

\subsection{Anchor admission, canonicalization, and frame extraction}
\label{app:method-graph}

This part details lines~\ref{alg:extract} to~\ref{alg:upsert} of
\Cref{alg:mirix}, which turn a committed episodic or semantic record into the
anchors and frames of \Cref{sec:hypergraph}.

\paragraph{Anchor keys and admission.}
The key \(\kappa(n)\) of an anchor name normalizes case, punctuation, articles
and other function words, word order, and conservative singular and plural
variants while keeping numbers. In the code it lowercases the name, replaces
punctuation with spaces, drops a fixed list of eleven articles and function
words, singularizes tokens of at least four letters, and sorts the remaining
tokens, so that \emph{Mobile App} and \emph{Mobile Apps} share one key. The key
is deliberately conservative. Numbers and dates are preserved, so \emph{10 Gallons} and \emph{20 Gallons}
stay distinct, and only plural endings are normalized, so \emph{Training} and
\emph{Trains} stay apart. A
single admission gate rejects generic filing terms, such as \emph{advice},
\emph{details}, \emph{information}, \emph{plan}, and \emph{tips}, and
dialogue-role pseudo-entities such as \emph{user} and \emph{assistant}, and it
keeps at most 16 anchors for an episodic record and 18 for a semantic one. The
caps bound the anchor space of a user, and they are sized for the arguments of
an assertion rather than for a bag of keywords, so that one frame with seven
arguments does not exhaust them. The gate is the \textsc{Admit} call of
\Cref{alg:mirix}, and because it runs on the canonical key, one entry such as
\emph{user} also blocks \emph{User}, \emph{Users}, and \emph{the user}.

\paragraph{Registry-guided canonicalization.}
For every extracted entity, approximate nearest-neighbor search retrieves up to four existing anchors of the user whose cosine
similarity is at least \(0.72\), and one language-model call per record makes a
conservative same-entity decision for all of its entities. The vector index
scores a pair by \((1+\cos)/2\), so the thresholds in the code are 0.86 here
and 0.93 for AutoDream (\Cref{app:method-autodream}). The call is instructed
not to merge different quantities, different days or months, different form or
model codes, or related but distinct concepts. Canonicalization runs before the admission gate, which is the order \(\textsc{Admit}(\textsc{Canon}^\star(\Phi))\) of
\Cref{alg:mirix}, so the gate judges canonical names rather than surface forms.
The candidates come from the anchors the user already has, which is what keeps
naming consistent as the graph grows and why we call the step registry-guided.

\paragraph{Frames, literals, and repair.}
The extractor behind \eqref{eq:frame} is instructed to keep dates, prices,
counts, durations, and measurements as literals on the frame rather than as
anchors, so that values such as 7, \$420, or three months stay attached to the
assertion they qualify. A frame is materialized only if its arguments resolve
to at least two distinct anchors of the user, which is the test \(|A_f|\ge 2\)
of \Cref{alg:mirix}. When a deterministic audit finds that the extraction
misses high-signal source details, such as quoted phrases, numbers, or concrete
visual content, one bounded repair call may add frames that the source
supports. The repair cannot remove anything the primary extraction produced.
The audit compares the words of the rendered frames with the source text. It
lists the quoted phrases and the numbers of the source that the frames do not
cover, it treats a source with visual content as uncovered when the frames
cover less than 45 percent of its content words, and it sends at most twelve
missing items to the repair call. A missing date alone does not trigger a
repair, because session dates recur in many records and are already preserved
by provenance. A missing date is added to a call that a missing quote, number,
or visual detail already requires. Repair is the \(\textsc{Repair}^\star\) call of
line~\ref{alg:extract}, and the frames it adds pass through the same
canonicalization, admission, and identity rules as the primary ones.

\subsection{Retrieval budgets, lanes, scores, and fusion}
\label{app:method-retrieval}

This part details the \textsc{Read} procedure of \Cref{alg:mirix}, whose steps
\Cref{sec:retrieval} introduces.

\begin{wraptable}{R}{0.36\textwidth}
\vspace{-0.6\baselineskip}
\centering
\captionsetup{justification=centering}
\caption{Row budgets \(B_q^B\) and \(B_q^R\) per query operator.}
\label{tab:lane-budgets}
\begin{tabular}{lrr}
\toprule
Operator & \(B_q^B\) & \(B_q^R\) \\
\midrule
Single & 6 & 4 \\
Temporal & 6 & 6 \\
Intersection & 7 & 7 \\
List-union & 7 & 9 \\
Count & 8 & 12 \\
\bottomrule
\end{tabular}
\vspace{-0.4\baselineskip}
\end{wraptable}
\paragraph{Operator budgets.}
The planner \eqref{eq:query-plan} fixes two row budgets for every query, and
\Cref{tab:lane-budgets} lists them by operator.
The budgets grow with the operator, because counting and set construction need
broader evidence than a single-value lookup. A single-value question needs the
one row that holds the value, so its cited-row budget is the smallest, whereas a count has to see many members of the set it counts. The evidence a
question inspects is therefore bounded by its operator and not by the length of
the history.

\paragraph{Anchor entry.}
Beyond the fifteen anchors that vector recall returns, a relation query adds up
to sixteen anchors that the question names exactly before the union is scored
by \eqref{eq:anchor-score}. Person hubs receive a larger penalty unless a
relation query names them explicitly. Only the exactly named anchors carry degree
and predicate information, so the hub and predicate terms act on them alone.
The two entry routes serve different questions. Vector recall finds the anchors
whose names paraphrase the question, and exact lookup finds the anchors that a
relation question names, so that the relation lane can start from the named
participants rather than from their nearest paraphrases.

\paragraph{Two evidence lanes.}
The lanes follow three rules beyond those of \Cref{sec:retrieval}. In the
base-recall lane, same-chunk support and immediate temporal neighbors are
recomputed from the provenance stored in the vector database instead of being duplicated as graph
bookkeeping edges. In the relation-citation lane, predicate filtering runs
inside Neo4j before the cap of 240 frames, and when the question names a
concrete object, only frames that contain it are kept. The base lane is the \(\textsc{Rows}(A_q)\) term and the relation lane the
\(\textsc{Adj}(A_q,\mathcal P_q)\) term of \Cref{alg:mirix}. Recomputing
same-chunk and temporal neighbors from provenance keeps the graph free of edges
that only restate what the evidence plane already records. Filtering by
predicate inside the database before the cap means that an explicitly named hub
with hundreds of incident frames does not trigger a full traversal, and the
frames that survive the cap are selected by a deterministic order on the frame
identifier, so which frames pass does not depend on the traversal order of the
database.

\paragraph{Fact scoring.}
Two details of \eqref{eq:fact-score} complete \Cref{sec:retrieval}. When a
frame has no extracted event time, the state term \(s_{\mathrm{state}}\) falls
back at half weight to the time of the source record, so the frame can still
compete on time. For intersection questions, a hard shared-object group is preserved before
these soft refinements, so recency cannot replace evidence that satisfies only
one side of the intersection. The role term reads the typed roles of a frame,
and the state term reads the status literal that marks an event as planned or
completed (\Cref{sec:hypergraph}), which is how planned events are penalized in count,
list-union, and temporal questions.

\paragraph{Citation-balanced fusion.}
Rows in each lane are reranked with a shallow vector-rank prior and the lexical
coverage of the query. Fusion runs separately for episodic and semantic rows
within the row limit of a request. Near-duplicate episodic paraphrases
are collapsed only when they share a calendar date, so the same wording on two
different dates stays separate. Cited rows are filled first because a graph
fact without the row that supports it would leave the answer unverifiable, and
the date condition on paraphrase collapse is the retrieval-side counterpart of
the occurrence identity \eqref{eq:episodic-frame-id}, which keeps undated
repetitions apart at write time.

\subsection{Answer-time verification}
\label{app:method-verification}

The \textsc{Answer} procedure of \Cref{alg:mirix} belongs to the evaluation
harness rather than to the memory server, and the rule-based check of
\Cref{sec:retrieval} never inspects a reference answer. The check fires when it
finds explicit uncertainty, a count without a number or one that disagrees with
a listed count, a temporal answer without a time, an answer number or month that
is absent from the evidence ledger and the question, ledger evidence with an
incompatible role, or a planned event in a count question. It also fires on
every intersection question. The extra search uses a query derived from
\(\Pi(q)\), or the question itself for a relation query. Most triggers mark a
way in which a draft falls short of the evidence it was given or disagrees with
it. A number or month that appears in neither the ledger nor the question was
not retrieved, a count that disagrees with a listed count contradicts the
evidence, and a planned event in a count question would count something that has
not happened. Intersection questions always trigger the check because they need
evidence for both sides, and a first retrieval that satisfies one side can look
complete. The rewrite is line~\ref{alg:retry} of \Cref{alg:mirix}, so the
procedure adds at most one search and one model call to a question.

\subsection{AutoDream schedule, candidates, and procedural refinement}
\label{app:method-autodream}

This part details the \textsc{SemanticDream} and \textsc{ProceduralDream}
procedures of \Cref{alg:mirix} (lines~\ref{alg:dream-pairs} to~\ref{alg:procedural}) and expands \Cref{sec:autodream}.

\paragraph{Invocation protocol.}
Semantic AutoDream is an explicit service call that never schedules itself. In
our protocol, the ingestion driver invokes AutoDream after every five chunks and
once after the final chunk. The chunk identifiers of each call resolve to the
newly written semantic rows, which define the frontier \(\Delta_u\) of
\eqref{eq:dream-contract}, and the eligible anchors backed by those rows are the
sources \(\mathcal A_\Delta\) of \eqref{eq:dream-candidates}. An intermediate
call therefore examines only the anchors whose semantic rows the batch just
wrote, and the final call also considers the eligible anchors that no earlier
call used as a source.

\paragraph{Candidate generation.}
The identity-stable types \(\mathcal T_{\mathrm{safe}}\) of
\Cref{sec:autodream} are person, organization, location, object, and concept.
Anchors of the types event, date, content, method, and \emph{other} are
excluded deliberately, because merging two of them can collapse distinct occurrences or temporal
context even when their names are close in embedding space. Both endpoints of a candidate pair must be backed by semantic rows and
share the same eligible type, and the cosine threshold \(0.86\) of
\eqref{eq:dream-candidates} is the index score \(0.93\) under the
\((1+\cos)/2\) scale of \Cref{app:method-graph}.

\paragraph{Bounded schedule.}
An intermediate Dream verifies at most 80 unseen pairs from the current semantic
delta. The final Dream considers the delta together with eligible anchors that
have never served as a Dream source, and it verifies at most 120 pairs. If that
round merges anything, one repair round examines only the touched
representatives and verifies at most 60 more pairs. Explicit negative decisions
are cached by anchor type, so later cycles do not
pay again to reconsider the same pair. The cache key hashes the case-folded
pair in sorted order together with the type, so the
order in which a pair is encountered does not matter. Only explicit rejections are cached. A pair that a round did not reach within
its cap is not recorded as rejected, so it becomes a candidate again whenever one of
its anchors is in a later frontier.
The intermediate rounds keep the graph consolidated while ingestion proceeds,
and the final round sweeps up anchors that earlier rounds never used as a source. In \Cref{alg:mirix}, the cache is the set
\(\mathcal N\) that line~\ref{alg:dream-pairs} subtracts from the candidate
pairs, and the per-round caps bound the number of \(\textsc{Same}^\star\) calls
the loop makes.

\paragraph{Procedural refinement.}
Two rules of the procedural path complete \Cref{sec:autodream}. The distiller
judges the outcome of each pending session from quotes in its transcript, and an
outcome whose quotes do not occur in the transcript is downgraded to unclear
rather than trusted. Besides creating, editing, or deleting a skill, the curator
can mark an experience as already covered, not worth a skill, or deferred, so it
can decline an experience without changing the store. The quote check on
outcomes and the quote check on citations are the procedural counterpart of the
provenance that the hypergraph keeps for facts. They keep the skill library
traceable although both \(\textsc{Distill}^\star\) and \(\textsc{Curate}^\star\)
in \Cref{alg:mirix} are language-model calls.

\subsection{Invariants, learned decisions, and complexity}
\label{app:method-invariants}\label{app:method-complexity}

\paragraph{Hard invariants.}
Four properties hold regardless of model behavior.
\begin{enumerate}[leftmargin=*,nosep]
    \item Every returned graph fact exposes the identifiers of its source
    memories.
    \item Undated episodic occurrences from different memory rows receive
    different frame identities.
    \item Retrieval and Dream both run under explicit candidate and output caps.
    \item A failed graph write cannot destroy a committed memory, and a graph
    read that raises an error falls back to direct evidence retrieval.
\end{enumerate}
Together with \Cref{eq:dream-contract}, the first and fourth invariants give
the source fidelity required in \Cref{sec:method-setup}, the third gives
bounded access, and the second gives temporal identity.

\paragraph{Learned and deterministic decisions.}
Language models route inputs, write typed memories, extract frames, repair
missing source details, judge semantic identity, and distill and curate
procedural experience. Together with answer generation, these are the calls
marked \(\star\) in \Cref{alg:mirix}. Deterministic code owns the
safety-critical boundaries, namely admissible anchor names, candidate
allowlists, frame identities, the representation of events and their status,
provenance attachment, query operators, lane budgets, merge algebra, and
mutation order. An LLM may propose that two anchors are identical, for instance,
but it cannot name an anchor outside the retrieved candidate set or delete
evidence directly.

\paragraph{Complexity.}

Let \(d\) be the number of admitted anchors in one memory, \(h\) the number of
frames extracted from it, \(K_A\) the anchor-retrieval cap, and \(K_F\) the
adjacent-frame cap. Ignoring model inference and indexed database lookups,
writing a memory into the graph costs \(O(d+h)\). Post-retrieval ranking and
fusion cost \(O(K_A+K_F+K_M)\), where \(K_M\) is the operator-dependent cited-row
budget, and the system never materializes an unbounded evidence set from the vector database.
For a Dream frontier of size \(|\Delta_u|\), nearest-neighbor generation costs
\(O(|\Delta_u|\log|\mathcal A_u|)\) under the vector index, and LLM verification
is capped at 80 pairs per intermediate cycle and 180 pairs across the two
possible final rounds. The constants behind these symbols are the ones of this
appendix. The admission gate bounds \(d\) by 16 for an episodic record and 18
for a semantic one (\Cref{app:method-graph}), \(K_A\) is the fifteen anchors kept after scoring, \(K_F\) is the cap of 240 frames, and \(K_M\) is \(B_q^R\), at most 12
(\Cref{app:method-retrieval}). The 180 final pairs are the 120 of the final
round plus the 60 of its repair round (\Cref{app:method-autodream}).

\subsection{Implementation and an example hypergraph}
\label{app:method-implementation}

\begin{figure}[H]
\centering
\includegraphics[width=\textwidth]{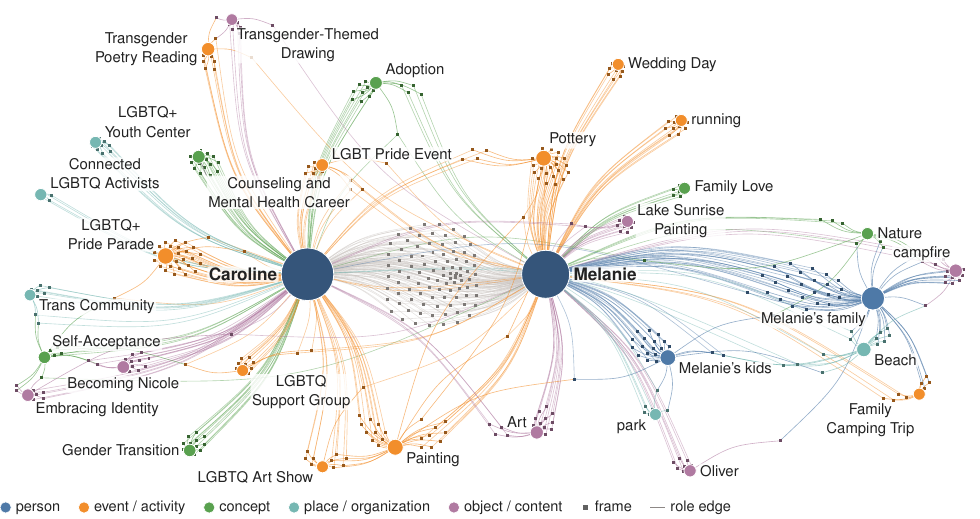}
\caption{The 32 anchors with the most frames in the conv-26 hypergraph and the 448 frames that join them. Anchors are colored by type and sized by frame count, and frames take the color of their nearest topic.}
\label{fig:graph-conv26}
\end{figure}

Evidence rows are stored in a vector database and the hypergraph in Neo4j.
\Cref{fig:graph-conv26} shows part of the hypergraph that \ours builds for
LoCoMo conversation conv-26, in which Caroline and Melanie talk from May to
October 2023. The full graph has 792 anchors, 1{,}379 frames, and 3{,}303 role
edges. The figure keeps the 32 anchors that appear in the most frames, leaving
out generic duplicates such as Family and kids, together with the 448 frames
that join at least two of them.

The two speakers are the largest hubs, with 650 and 529 frames. Topics that
one speaker raises gather on that speaker's side, such as the pride parade and
\emph{Becoming Nicole} for Caroline and pottery and her family for Melanie. The
94 frames that join only the two speakers lie between the hubs and are drawn in
gray, and painting and art are reached from both sides. Each frame also keeps role labels,
literals such as dates, and citations to its source records, which the figure
leaves out. Through these frames one anchor gathers evidence across sessions.
Self-Acceptance appears in 13 frames from 10 sessions, which link it to the
LGBTQ support group, the pride parade, \emph{Becoming Nicole}, and Caroline's
painting \emph{Embracing Identity}.

We built this graph with graph indexing and AutoDream enabled, using Claude
Haiku 5.5 for extraction and Cohere Embed v4 for embeddings instead of the
models described in \Cref{sec:exp-setup}.

\section{Additional results}
\label{app:results}

This appendix gives the full tables behind \Cref{fig:res-ablation,fig:res-baselines,fig:res-agent,fig:res-alfworld-types,fig:res-cost}.
Accuracy and success rates are percentages, and differences are in points.

\subsection{User memory}
\label{app:results-user}

\Cref{tab:lme} lists the LongMemEval-S accuracy of every system, judged like LoCoMo
by \texttt{gpt-4o-mini} with majority-of-3 voting. It reports M0, M1, and M2, so it complements the LoCoMo
ablation of \Cref{tab:locomo}. \Cref{tab:docqa} gives the results on the three document QA tasks of
MemoryAgentBench. SHDocQA and MHDocQA ask single-hop and multi-hop questions about
long documents and use substring matching on 100 questions each, and DetectiveQA uses
letter exact matching on 71 questions. The document QA runs use M1, with AutoDream disabled by experimental choice.

We ran HyperMem ourselves with its original settings unchanged, and all HyperMem
results in this paper come from these runs. On LoCoMo, our run reaches 91.0\%,
below the 92.7\% reported by \citet{hypermem}.

\suppressfloats[t]
\begin{table}[t]
\centering
\begin{minipage}[t]{0.43\textwidth}
\centering
\caption{LongMemEval-S accuracy (\%) on the evaluated 300 questions.}
\label{tab:lme}
\small
\begin{tabular}{lc}
\toprule
Method & Accuracy \\
\midrule
mem0        & 56.3 \\
A-Mem       & 45.0 \\
MemoryBank  & 39.7 \\
Graphiti    & 23.0 \\
HyperMem    & 58.3 \\
\midrule
MIRIX-1.0 (M0)              & 58.8 \\
\ours\ (M1, graph)           & {70.0} \\
\ours\ (M2, graph $+$ dream) & \textbf{72.6} \\
\bottomrule
\end{tabular}
\end{minipage}\hfill
\begin{minipage}[t]{0.53\textwidth}
\centering
\caption{Document QA accuracy (\%). \ours uses M1, with AutoDream disabled.}
\label{tab:docqa}
\small
\begin{tabular}{lccc}
\toprule
Method & SHDoc & MHDoc & Detective \\
\midrule
mem0        & 48.0 & 22.0 & 53.5 \\
A-Mem       & 14.0 & 20.0 & 62.0 \\
MemoryBank  & 34.0 & 28.0 & \textbf{67.6} \\
Graphiti    & 50.0 & 43.0 & 54.9 \\
HyperMem    & 52.0 & 49.0 & 66.2 \\
\ours\ & \textbf{83.0} & \textbf{71.0} & 64.8 \\
\bottomrule
\end{tabular}
\end{minipage}
\end{table}

On LongMemEval-S the graph increment of 11.2 points accounts for most of the
14.3-point margin over HyperMem. The
flat configuration M0 reaches 58.8\%, close to HyperMem at 58.3\% and mem0 at 56.3\%,
while M1 reaches 70.0\% and M2 72.6\%, 14.3 points above HyperMem. On SHDocQA and MHDocQA, HyperMem is the strongest baseline, and M1 leads it by 31.0
and 22.0 points. DetectiveQA is the exception.
MemoryBank leads with 67.6\%, followed by HyperMem with 66.2\% and M1 with 64.8\%.
A-Mem varies the most across the three document tasks, from 14.0\% on SHDocQA to
62.0\% on DetectiveQA.

\Cref{tab:memcd-gen} completes the MemoryCD results of \Cref{tab:memcd-summ} with
review generation for the same 100 users in the Home\_and\_Kitchen domain. In both
MemoryCD tables, cross-domain memory comes from the other three Amazon domains, and
single-domain memory and testing use the same domain, so the cross-domain setting asks
whether preferences stored for other products carry over to a new domain. R-1, R-L,
B-4, and BERT denote ROUGE-1, ROUGE-L, BLEU-4, and BERTScore-F1, higher is better, and
bold marks column maxima, including ties.

\begin{table}[H]
\centering
\caption{Review generation on MemoryCD, with the users and domains of \Cref{tab:memcd-summ}.}
\label{tab:memcd-gen}
\small
\begin{tabular}{lcccc|cccc}
\toprule
& \multicolumn{4}{c|}{Cross-domain} & \multicolumn{4}{c}{Single-domain} \\
\cmidrule(lr){2-5}\cmidrule(lr){6-9}
Method & R-1 & R-L & B-4 & BERT & R-1 & R-L & B-4 & BERT \\
\midrule
Long-context & .2842 & .1434 & .0092 & .8521 & .2865 & .1455 & .0105 & .8520 \\
RAG          & .2727 & .1354 & .0070 & .8484 & .2818 & .1407 & .0078 & .8498 \\
mem0         & .2783 & .1432 & .0087 & .8514 & .2780 & .1391 & .0083 & .8502 \\
A-Mem        & .2836 & .1434 & .0094 & .8520 & .2857 & .1445 & .0101 & .8517 \\
MemoryBank   & .2799 & .1424 & .0073 & .8505 & .2835 & .1424 & .0089 & .8514 \\
Graphiti     & .2717 & .1379 & .0083 & .8496 & .2808 & .1441 & .0099 & .8512 \\
\ours\  & \textbf{.2846} & \textbf{.1491} & \textbf{.0108} & \textbf{.8544} & \textbf{.2868} & \textbf{.1487} & \textbf{.0106} & \textbf{.8539} \\
\bottomrule
\end{tabular}
\end{table}

\ours has the highest value in every column of both settings, but the margins are
small. In the cross-domain setting its ROUGE-L of .1491 is .0057 above the next best,
.1434 for Long-context and A-Mem, and its BLEU-4 of .0108 is .0014 above A-Mem.
BERTScore spans at most .006 across all systems in either setting. In the
single-domain setting the margins shrink further, for example .2868 against .2865
for Long-context on ROUGE-1. 

\subsection{Agent memory}
\label{app:results-agent}

\Cref{tab:alfworld-proc} breaks down the ALFWorld success rates of \Cref{fig:res-agent}
by task type for both backbones, with the number of held-out episodes of each type in
parentheses, 134 in total. \textbf{All} is the overall success rate, given as the mean
and the sample standard deviation over three runs. Average steps are taken over all
episodes, with a cap of 50 steps, and over successful episodes only, and lower is
better. For each model, bold marks the best of the three memory methods.
\Cref{fig:res-alfworld-types} plots the GPT-6 Luna rows.

\suppressfloats[t]
\begin{table}[t]
\centering
\caption{ALFWorld success by task type (episodes per type in parentheses) and average steps, over three runs.}
\label{tab:alfworld-proc}
\small
\setlength{\tabcolsep}{3.3pt}
\begin{adjustbox}{max width=\textwidth}
\begin{tabular}{llccccccccc}
\toprule
 & & \multicolumn{7}{c}{Success rate (\%)} & \multicolumn{2}{c}{Avg.\ steps} \\
\cmidrule(lr){3-9} \cmidrule(lr){10-11}
Model & Method & Pick (24) & Look (18) & Clean (31) & Heat (23) & Cool (21) & Pick Two (17) & All & all & succ. \\
\midrule
\multirow{4}{*}{GPT-6 Luna}
  & No memory     & 83.3 & 85.2 & 60.2 & 47.8 & 81.0 & 64.7 & 69.4$\pm$0.7 & 25.2 & 14.2 \\
  & ReasoningBank & 70.8 & 66.7 & 65.6 & 63.8 & 82.5 & 49.0 & 66.9$\pm$2.8 & 26.4 & 14.7 \\
  & AWM           & 77.8 & 77.8 & 64.5 & \textbf{75.4} & \textbf{88.9} & 58.8 & 73.6$\pm$1.9 & 23.2 & 13.6 \\
  & \ours         & \textbf{79.2} & \textbf{83.3} & \textbf{74.2} & 73.9 & 85.7 & \textbf{60.8} & \textbf{76.4$\pm$6.2} & \textbf{21.9} & \textbf{13.2} \\
\midrule
\multirow{4}{*}{GPT-5.6 Luna}
  & No memory     & 73.6 & 63.0 & 37.6 & 50.7 & 66.7 & 37.3 & 54.2$\pm$1.1 & 31.6 & 16.1 \\
  & ReasoningBank & \textbf{75.0} & 50.0 & 50.5 & 53.6 & 46.0 & 31.4 & 52.2$\pm$0.7 & 31.4 & 14.4 \\
  & AWM           & 69.4 & 61.1 & 44.1 & \textbf{63.8} & 65.1 & \textbf{47.1} & 58.0$\pm$2.8 & 29.9 & 15.3 \\
  & \ours         & 69.4 & \textbf{75.9} & \textbf{54.8} & \textbf{63.8} & \textbf{66.7} & 39.2 & \textbf{61.7$\pm$14.0} & \textbf{26.8} & \textbf{12.4} \\
\bottomrule
\end{tabular}
\end{adjustbox}
\end{table}

With GPT-6 Luna, \ours is the best of the three memory methods on four of the six
task types, Pick, Look, Clean, and Pick Two, and AWM is ahead on Heat and Cool.
Against no memory, Clean gains 14.0 points, Heat 26.1, and Cool 4.7, while Pick,
Look, and Pick Two lose 4.1, 1.9, and 3.9 points.
With GPT-5.6 Luna, \ours leads the memory methods on Look, Clean, and Cool and ties
AWM on Heat at 63.8\%, while ReasoningBank is ahead on Pick and AWM on Pick Two. With
both backbones \ours takes the fewest steps, 21.9 and 26.8 over all episodes and 13.2
and 12.4 over successful ones, but its overall standard deviation of 14.0 points with
GPT-5.6 Luna is the largest in the table.

\Cref{tab:webshop-proc} gives the same breakdown for WebShop with GPT-6 Luna over the
five product categories of the 140 test instructions, with a cap of 15 steps. A
WebShop episode succeeds when its reward is 1, and bold again marks the best of the
three memory methods.

\begin{table}[H]
\centering
\caption{WebShop success by product category and average steps with GPT-6 Luna, over three runs.}
\label{tab:webshop-proc}
\small
\setlength{\tabcolsep}{3.3pt}
\begin{adjustbox}{max width=\textwidth}
\begin{tabular}{lcccccccc}
\toprule
& \multicolumn{6}{c}{Success rate (\%)} & \multicolumn{2}{c}{Avg.\ steps} \\
\cmidrule(lr){2-7} \cmidrule(lr){8-9}
Method & Fashion (30) & Beauty (21) & Electronics (29) & Garden (32) & Grocery (28) & All & all & succ. \\
\midrule
No memory & 54.4 & 31.7 & 21.8 & 19.8 & 32.1 & 31.9$\pm$1.8 & 7.9 & 6.2 \\
ReasoningBank & 43.3 & 33.3 & 12.6 & 17.7 & 23.8 & 25.7$\pm$3.8 & 10.2 & 6.9 \\
AWM & 53.3 & 31.7 & \textbf{28.7} & \textbf{27.1} & \textbf{38.1} & 36.0$\pm$4.3 & 9.1 & 6.9 \\
\ours & \textbf{57.8} & \textbf{39.7} & 25.3 & \textbf{27.1} & \textbf{38.1} & \textbf{37.4$\pm$2.1} & \textbf{8.2} & \textbf{6.3} \\
\bottomrule
\end{tabular}
\end{adjustbox}
\end{table}

\ours has the highest overall success, 37.4\% against 36.0\% for AWM and 31.9\%
without memory, and the smallest spread among the memory methods, 2.1 points. Its margins over AWM are 4.5 points on Fashion and 8.0 on Beauty. It matches AWM on
Garden and Grocery, at 27.1\% and 38.1\%, and trails it by 3.4 points on Electronics. ReasoningBank falls below no memory overall, as it does on ALFWorld with both
backbones, and on four of the five categories. All three memory methods take more
steps over all episodes than no memory, 8.2 to 10.2 against 7.9, and \ours takes the
fewest of the three.

\subsection{Costs}
\label{app:results-cost}

\Cref{tab:eff-docqa} gives the measurements behind \Cref{fig:res-cost}. Ingest is the
total wall-clock time in seconds, and Ans./q is the mean time to retrieve and answer
one question. Answer tokens sum input and output over all questions, where K and M
denote thousands and millions. Graph size counts Neo4j nodes and edges and excludes
the evidence store. M0 builds no graph, and for HyperMem the column names its index type instead of a size. None of these measurements includes
AutoDream, so they characterize M1 and not the full M2 system.

\suppressfloats[t]
\begin{table}[t]
\centering
\caption{Cost of memory for HyperMem, M0, and M1, all without AutoDream.}
\label{tab:eff-docqa}\label{tab:eff-conv}
\small
\begin{tabular}{llcccc}
\toprule
Benchmark & Method & Ingest (s) & Ans./q (s) & Ans.\ tok. & Graph (nodes/edges) \\
\midrule
\multirow{3}{*}{SHDoc}
 & HyperMem       & 5{,}665 & 10.55 & 3.79M & hypergraph \\
 & MIRIX-1.0 (M0) & 6{,}227 & 3.76 & 763K & none \\
 & \ours\ (M1)    & 23{,}784 & 5.45 & 1.40M & 29{,}341 / 33{,}526 \\
\midrule
\multirow{3}{*}{MHDoc}
 & HyperMem       & 11{,}031 & 10.65 & 3.06M & hypergraph \\
 & MIRIX-1.0 (M0) & 13{,}792 & 7.46 & 1.23M & none \\
 & \ours\ (M1)    & 55{,}301 & 9.01 & 1.33M & 49{,}046 / 55{,}288 \\
\midrule
\multirow{3}{*}{Detective}
 & HyperMem       & 11{,}653 & 13.05 & 0.35M & hypergraph \\
 & MIRIX-1.0 (M0) & 10{,}321 & 5.11 & 706K & none \\
 & \ours\ (M1)    & 39{,}491 & 7.24 & 1.16M & 36{,}987 / 46{,}651 \\
\midrule
\multirow{2}{*}{LoCoMo}
 & HyperMem     & 69{,}467 &  9.70 & 16.5M & hypergraph \\
 & \ours\ (M1)  & 21{,}416 &  5.46 & 15.7M & 23{,}820 / 32{,}394 \\
\midrule
\multirow{2}{*}{LongMemEval-S}
 & HyperMem     & 170{,}224 & 10.73 & 4.10M & hypergraph \\
 & \ours\ (M1)  & 70{,}626 & 10.10 & 4.02M & 181{,}131 / 205{,}428 \\
\bottomrule
\end{tabular}
\end{table}

\Cref{tab:eff-docqa} shows where graph construction costs time. On the three document tasks, M1
ingests 3.8 to 4.0 times more slowly than M0 and 3.4 to 5.0 times more slowly than
HyperMem. On the conversational benchmarks the order reverses, and M1 needs 21{,}416\,s
for LoCoMo against 69{,}467\,s for HyperMem and 70{,}626\,s for LongMemEval-S against
170{,}224\,s. Answering is faster for M1 than for HyperMem on all five benchmarks, by between 0.63\,s per question on LongMemEval-S and 5.81\,s on DetectiveQA, although M0 answers
the document questions faster still. M1 uses fewer answer tokens than HyperMem on four
benchmarks and more on DetectiveQA, 1.16M against 0.35M. Its graphs range from 23{,}820
nodes and 32{,}394 edges for LoCoMo to 181{,}131 nodes and 205{,}428 edges for
LongMemEval-S.

For \Cref{tab:agent-token-consumption}, training covers actor interactions and memory
learning, and testing covers actor interactions and memory retrieval on all test
tasks, including unsuccessful ones. Counts include LLM input and output tokens, with
reasoning tokens in the output and cached tokens in the input, and exclude embedding
tokens. No memory skips training.

\end{document}